\documentclass[journal]{IEEEtran}

\usepackage{graphicx,epsfig}
\usepackage{amsmath,mathrsfs} 
\usepackage{algorithm}
\usepackage[noend]{algpseudocode}
\makeatletter
\def\BState{\State\hskip-\ALG@thistlm}
\makeatother
\usepackage{amssymb} 
\usepackage{amsthm}
\usepackage{amsfonts}
\usepackage{mathtools}
\usepackage{multirow}
\usepackage{mathtools} 
\usepackage{cite}
\usepackage{epstopdf}
\usepackage{ifthen}
\usepackage{bbm}
\usepackage{setspace}

\newcommand{\C}{ \mathbb{C}}
\newcommand{\R}{ \mathbb{R}}

\newcommand{\N}{ \mathbb{N}}

\newcommand{\eqdef}{\stackrel{\vartriangle}{=}}

\newcommand{\ud}{\,\mathrm{d}}

\def\V#1{{\boldsymbol{#1}}}         
\def\Spc#1{{\mathcal{#1}}}  
\def\M#1{{\bf{#1}}}  
\def\Op#1{{\mathrm{#1}}}  

\makeatletter
\def\blfootnote{\xdef\@thefnmark{}\@footnotetext}
\makeatother

\DeclareMathOperator*{\argmin}{arg\,min}

\usepackage{tikz,pgfplots}
\pgfplotsset{compat=1.5.1}
\usepgfplotslibrary{groupplots}
\usetikzlibrary{intersections,calc,arrows,matrix,spy,math,calc,fit,positioning, decorations.pathmorphing}
\tikzset{snake it/.style={decorate, decoration=snake}}

\usetikzlibrary{arrows.meta,shapes.geometric}

\usepackage{ dsfont,color }
\usepackage[nottoc, notlof, notlot]{tocbibind}
\usepackage{bbm}
\usepackage{stackengine}
\usepackage{graphicx}
\usepackage{siunitx}
\usepackage{subcaption}

\begin{document}
\IEEEoverridecommandlockouts

\title{Bayesian Inversion for Nonlinear Imaging Models Using Deep Generative Priors}

\author{Pakshal~Bohra, Thanh-an~Pham, Jonathan~Dong, and~Michael~Unser,~\IEEEmembership{Fellow,~IEEE}
\thanks{This work was supported in part by the Swiss National Science Foundation under Grant 200020\_184646 / 1 and in part by the European Research Council (ERC Project FunLearn) under Grant 101020573. 

Pakshal Bohra, Jonathan Dong and Michael Unser are with the Biomedical Imaging Group, \'{E}cole polytechnique f\'{e}d\'{e}rale de Lausanne, 1015 Lausanne, Switzerland (e-mail: pakshal.bohra@epfl.ch; jonathan.dong@epfl.ch; michael.unser@epfl.ch).

Thanh-an Pham was with the Biomedical Imaging Group, École polytechnique fédérale de Lausanne, Lausanne, Switzerland. He is now with the 3D Optical Systems Group, Department of Mechanical Engineering, Massachusetts Institute of Technology, 77 Massachusetts Ave, Cambridge, Massachusetts 02139, USA (e-mail: tampham@mit.edu).
}
}

\maketitle

\begin{abstract}
  Most modern imaging systems incorporate a computational pipeline to infer the image of interest from acquired measurements. The Bayesian approach to solve such ill-posed inverse problems involves the characterization of the posterior distribution of the image. It depends on the model of the imaging system and on prior knowledge on the image of interest. In this work, we present a Bayesian reconstruction framework for nonlinear imaging models where we specify the prior knowledge on the image through a deep generative model. We develop a tractable posterior-sampling scheme based on the Metropolis-adjusted Langevin algorithm for the class of nonlinear inverse problems where the forward model has a neural-network-like structure. This class includes most practical imaging modalities. We introduce the notion of augmented deep generative priors in order to suitably handle the recovery of quantitative images. We illustrate the advantages of our framework by applying it to two nonlinear imaging modalities---phase retrieval and optical diffraction tomography.

\end{abstract}


\section{Introduction}
In practical imaging systems, the object of interest $\M s \in \R^K$ is observed indirectly by performing a series of measurements \mbox{$\M y \in \C^M$}. Mathematically, this process is often modeled as
\begin{equation}
  \M y = \M H (\M s) + \M n,
\end{equation}
where $\M H: \R^K \rightarrow \C^M$ is an operator that describes the physics of the imaging system and $\M n \in \C^M$ is an additive noise. The operator $\M H$ can be linear or nonlinear, depending on the imaging modality. For example, in magnetic resonance imaging, one captures noisy samples of the Fourier transform of the signal. 
The task at hand is then to reconstruct the signal $\M s$ from the obtained measurements $\M y$. Typically, such inverse problems are ill-posed, in the sense that there exist a multitude of signals which produce identical measurements. Thus, one cannot rely on direct inversion techniques to obtain relevant solutions. 

\subsection{Variational Methods}
In variational methods, the solution to the inverse problem is specified as the minimizer of a cost functional
\begin{equation}\label{eq:variational_formulation}
    \widehat{\M s} = \argmin_{\M s \in \R^K} \Big( E\big(\M y, \M H (\M s) \big) + \tau R(\M s) \Big),
\end{equation}
where the data-fidelity term $E: \C^{M} \times \C^{M} \rightarrow \R_{+}$ forces the solution to be consistent with the measurements, the regularization $R: \R^{K} \rightarrow \R_{+}$ imposes some prior constraints on the solution, and $\tau \in \R_{+}$ is a tunable hyperparameter. Typical candidates for these terms are $E\big(\M y, \M H (\M s) \big) = \|\M y - \M H (\M s)\|_{2}^{2}$ and $R(\M s) = \|\M L \M s\|_{p}^{p}$ \cite{tikhonov1963,bertero1998, rudin1992nonlinear, lustig2007, figueiredo2007, lim2015comparative, kamilov2016optical} with $p \in [1, 2]$. Here, $\M L$ is a linear transformation such as the discrete version of the wavelet transform or the gradient operator, which takes part in the regularization. For instance, total-variation (TV) regularization \cite{rudin1992nonlinear} uses the $\ell_1$-norm along with the gradient operator, which promotes solutions with sparse derivatives. It is widely used for compressed sensing and extreme imaging applications where the data is scarce \cite{kamilov2016optical}. The resulting optimization problems are typically solved by iterative algorithms such as gradient descent, the fast iterative shrinkage-thresholding algorithm (FISTA) \cite{figueiredo2003algorithm, daubechies2004iterative,beck2009fast}, or the alternating-direction method of multipliers (ADMM) \cite{boyd2011distributed}.

\subsection{Bayesian Inference}
In the Bayesian approach to image reconstruction \cite{mohammad1996full, mohammad2002bayesian, stuart2010inverse, dashti2013bayesian}, the signal $\M s$ is modeled as the realization of a random vector with a suitable probability density function (pdf) $p_{\mathrm{S}}$ that captures our prior knowledge about the signal. The idea here is to characterize the posterior distribution
\begin{equation}\label{eq:posterior_dist}
  p_{\mathrm{S}|\mathrm{Y}}(\M s | \M y) \propto p_{\mathrm{N}}\big(\M y - \M H (\M s)\big) p_{\mathrm{S}}(\M s),  
\end{equation}
which depends on the statistics of the noise $p_{\mathrm{N}}$ and on the prior distribution $p_{\mathrm{S}}$, and to make inferences based on it. 

The posterior distribution can be used for the derivation of several point estimators for the signal $\M s$. One such example is the maximum a posteriori estimator, which is the mode of the posterior distribution and leads to an optimization problem that resembles \eqref{eq:variational_formulation}, with $E\big(\M y, \M H (\M s)\big) \propto \big(-\log \big(p_{\mathrm{N}}\big(\M y - \M H (\M s)\big)\big)\big)$ and $R(\M s) \propto \big(-\log \big(p_{\mathrm{S}}(\M s)\big)\big)$, thus linking the variational and Bayesian approaches \cite{babacan2009bayesian, unser2010stochastic,gribonval2011should, pereyra2019revisiting}. Another example is the minimum mean-square error (MMSE) estimator which turns out to be the posterior mean \cite{pereyra2019revisiting}.

Besides the derivation of point estimators, the Bayesian framework allows one to quantify the uncertainty of the reconstructed image. This feature offers an interesting perspective for computational imaging as most practical reconstruction schemes, including the variational ones, do not provide any assessment of reliability. 

In general, inference tasks entail the estimation of expected values with respect to the posterior distribution. Typically, these are high-dimensional integrals that cannot be evaluated analytically. Thus, one relies on Markov chain Monte Carlo (MCMC) methods to efficiently draw samples from the posterior and then use them to approximate the integrals \cite{gilks1995markov, geyer1992practical, pereyra2015survey, kaipio2006statistical}.

\subsection{Deep-Learning-Based Methods}
Over the past few years, researchers have started to deploy deep-learning-based methods to solve inverse problems in imaging. The learning-based methods have been found to outperform the traditional model-based ones. Broadly speaking, their underlying principle is to utilize large amounts of training data to improve the reconstruction quality, as opposed to the specification of prior information about the image of interest in the form of mathematical models, as in the variational and Bayesian approaches described earlier.

The first generation of deep-learning-based methods involves training a convolutional neural network (CNN) as a nonlinear mapping that relates a low-quality estimate of the signal to the desired high-quality estimate \cite{jin2017deep,chen2017low,hyun2018deep,monakhova2019learned,perdios2021cnn}. The reconstruction pipeline then consists of using a fast classical algorithm to yield an initial solution and then correcting for its artifacts using the trained CNN. This category of methods includes ``unrolling" \cite{gregor2010learning,chen2016trainable,sun2016deep,aggarwal2018modl,adler2018learned,monga2021algorithm}, where the architecture of the CNN is designed by studying iterations of algorithms used for solving Problem \eqref{eq:variational_formulation}. While the first-generation end-to-end learning methods have achieved state-of-the-art performances in several inverse problems, recent works have highlighted their instability and lack of robustness \cite{antun2020instabilities,gottschling2020troublesome}.  

The second generation of deep-learning-based methods aims at the integration of CNNs into iterative reconstruction algorithms. The plug-and-play priors (PnP) \cite{venkatakrishnan2013plug} and regularization-by-denoising (RED) \cite{romano2017little} frameworks are two successful examples that provide a way to carry out this integration. In PnP algorithms, the proximal operator that appears in the iterations of the proximal algorithms (FISTA, ADMM) is replaced by a generic denoiser which imposes an implicit prior on the signal. RED, by contrast, incorporates an explicit regularization term that is constructed with the help of the chosen denoiser. In the learning-based variants of these frameworks, one uses appropriately trained CNNs as the denoising routines \cite{tirer2018image,ryu2019plug,sun2019block,wu2020simba,liu2020rare,zhang2021plug,sun2021scalable}. Another example of such methods is projected gradient descent where the projection operator is a trained neural network that projects onto the space of desired signals \cite{rick2017one,gupta2018cnn,yang2020deep}. Unlike the first-generation methods, the second-generation ones enforce consistency between the reconstructed signal and the measurements. They are also more versatile as the CNN denoisers can be used for several inverse problems without the need for retraining. One obstacle to the deployment of these learning-based iterative schemes is that the Lipschitz constant of the CNNs must be controlled in order to ensure their convergence \cite{bauschke2011convex,hertrich2020convolutional}, which is not straightforward and remains an active area of research \cite{hertrich2020convolutional,terris2020building,bohra2021learning}. 

One can also identify a third class of deep-learning-based methods that make use of deep generative models such as variational autoencoders (VAE) \cite{kingma2013auto} and generative adversarial networks (GAN) \cite{goodfellow2014generative}. These models include a generator network that maps a low-dimensional latent space to the high-dimensional signal space. They are trained to capture the statistics of the dataset and generate sample signals similar to those in the dataset. Once such a deep generative model has been successfully trained, its application to an inverse problem typically consists of finding the optimal latent variable such that the resulting signal best fits the measurements. Recent works have focused on the design and analysis of algorithms for the inversion of such generative models \cite{bora2017compressed,shah2018solving, hand2018phase, huang2021provably, gonzalez2022solving}.

The three classes of deep-learning-based methods discussed so far are variational in nature and provide a single reconstruction as their output. The success of these methods has stimulated the development of Bayesian methods that exploit the power of neural networks. For instance, in \cite{adler2018deep}, the authors propose two frameworks for ``deep Bayesian inversion" that are analogues of the first generation end-to-end deep-learning-based methods and require training data consisting of signals and their corresponding measurements. Their first approach involves the training of a conditional GAN to sample from the posterior distribution, while their second approach deploys neural networks to approximate a chosen statistical estimator. More recently, the focus has been on the development of more modular Bayesian methods where only the prior is modeled by neural networks. This has led to various posterior sampling schemes for priors defined either implicitly through denoising CNNs (such as the ones used in the PnP or RED frameworks) \cite{kadkhodaie2020solving,kawar2021snips,laumont2022bayesian} or through GANs \cite{patel2019bayesian}, VAEs \cite{tezcan2022sampling, holden2022bayesian}, and score-based generative models \cite{jalal2021robust,song2021solving}. So far, most of these works have focused on inverse problems with linear or linearized forward models. 

A current frontier in imaging is the inversion of nonlinear models, which arise in several applications, two notable examples being phase retrieval and optical diffraction tomography. Such applications could benefit greatly from the development of neural-network-based Bayesian reconstruction methods.

\subsection{Contributions}
In this paper, we present a Bayesian framework to solve a broad class of nonlinear inverse problems, where the prior is represented by a trained deep generative model. Our contributions are as follows.
\begin{itemize}
\item We develop a method based on the Metropolis-adjusted Langevin algorithm (MALA) \cite{roberts1996exponential, roberts2002langevin} to sample from the posterior distribution for the class of nonlinear inverse problems where the forward model has a neural-network-like structure. This class includes a wide variety of practical imaging modalities. We show that the structure of the forward model and the low-dimensional latent space of the generative prior enable tractable Bayesian inference.  
 
\item We introduce the concept of augmented generative models. This is motivated by the observation that deep generative models are easier to train when the dataset consists of images with the same range of pixel values. Unfortunately, such models are not well-matched to imaging modalities where one is interested in extracting the precise value of objects rather than merely visualizing contrast. Our proposed augmented models provide us with a simple but effective way of dealing with quantitative data.

\item We illustrate the advantages of the proposed reconstruction framework through numerical experiments for two nonlinear imaging modalities: phase retrieval and optical diffraction tomography.
\end{itemize}

The paper is organized as follows: In Section \ref{sec:forward_models}, we discuss the structure of the forward model for our nonlinear inverse problems. We detail the Bayesian reconstruction framework in Section \ref{sec:bayesian_framework}. There, we introduce augmented generative models and we explain our posterior-sampling scheme. We present our experimental results in Section \ref{sec:experiments}.

\section{Nonlinear Inverse Problems and Forward Models} \label{sec:forward_models}
In this section, we start by describing the class of nonlinear inverse problems that we are interested in. We then focus on two concrete examples---phase retrieval and optical diffraction tomography---and detail the physical models involved.

\subsection{Nonlinear Inverse Problems}\label{subsec:forward_model_structure}
The objective is to recover an image $\M s \in \R^K$ from its noisy measurements $\M y \in \C^M$ given by $\M y = \M N \big( \M y_0 \big)$ with
\begin{equation}\label{eq:nonlinear_forward_general}
 \M y_0 = \M H (\M s),
\end{equation}
where $\M H : \R^K \rightarrow \C^M$ is a nonlinear operator that models the physics of the imaging system  and $\M N : \C^M \rightarrow \C^M$ is an operator that models the corruption of the measurements by noise. In this work, we consider the class of nonlinear forward models $\M H$ whose computational structure can be encoded by a directed acyclic graph and thus resembles a neural network.

The Jacobian matrix of $\M H$ at any point $\M x = (x_1, \ldots, x_K) \in \R^K$ is defined as
\begin{equation}
  \M J_{\M H}(\M x) = \begin{bmatrix}
  \frac{\partial}{\partial x_1} [\M H(\M x)]_{1} & \cdots & \frac{\partial}{\partial x_K} [\M H(\M x)]_{1} \\
  \vdots & \ddots & \vdots \\
  \frac{\partial}{\partial x_1} [\M H(\M x)]_{M} & \cdots & \frac{\partial}{\partial x_K} [\M H(\M x)]_{M}
  \end{bmatrix}.
\end{equation}
Gradient-based MCMC methods (see Section \ref{sec:bayesian_framework} for a specific example) involve the computation of quantities such as ${\M J^{H}_{\M H}(\M x)} \M r$ for some vectors $\M x \in \R^K$, $\M r \in \C^M$, and this can be a potential bottleneck. The neural-network-like structure of $\M H$ allows us to compute these efficiently using the error backpropagation algorithm. This, in turn, makes Bayesian inference computationally feasible.   

The class of nonlinear inverse problems that fit this description is very broad and adaptable to most existing imaging modalities. In principle, it covers all possible inverse problems, in particular, the linear case is trivially covered. More generally, if sufficient data is available, one can indeed train a neural network to mimic the physics of our forward model. Next, we look at two particular problems that nicely fall within our predefined class.

\subsection{Phase Retrieval}

Phase retrieval \cite{shechtman2015phase,fogel2016phase} is a nonlinear inverse problem that is ubiquitous in computational imaging. It consists in the recovery of a signal from its intensity-only measurements and is a central issue in optics \cite{millane1990phase,maiden2009improved}, astronomy \cite{fienup1993hubble,freedman2001final}, and computational microscopy \cite{maleki1993phase,gureyev1997rapid,zernike1942phase,zheng2013wide}.

\begin{figure}[t]
  \centering
  \includegraphics[width=0.25\textwidth]{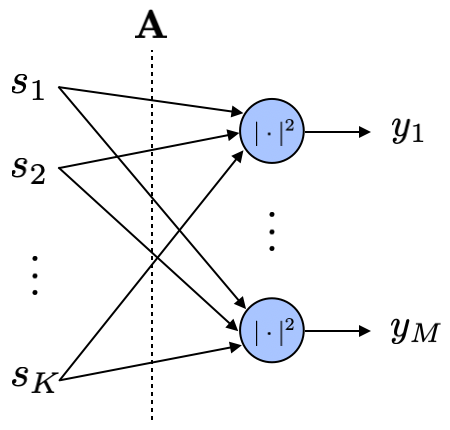}
 \caption{The forward model for phase retrieval \eqref{eq:phase_retrieval_forward} expressed as a one-layer fully-connected neural network with linear weights $\M A$ and quadratic activation functions.}
 \label{fig:phase_retrieval_forward}
\end{figure} 

In the phase-retrieval problem that we consider in this paper, the noise-free measurements are modeled as
\begin{equation}\label{eq:phase_retrieval_forward}
  \M y_0 = \M H_{\text{pr}} (\M s) = |\M A \M s|^2,
\end{equation}
where $\M A: \R^K \rightarrow \C^M$ is either the Fourier matrix \cite{millane1990phase, zheng2013wide,rodenburg2004phase} or some realization of a random matrix with independent and identically distributed (i.i.d.) elements \cite{candes2015phase,mondelli2018fundamental,fogel2016phase}, and where $|\cdot|^2$ is a component-wise operator. As shown in Figure \ref{fig:phase_retrieval_forward}, the forward model in \eqref{eq:phase_retrieval_forward} can be expressed as a one-layer fully-connected neural network with fixed linear weights $\M A$ and quadratic activation functions.

\subsection{Optical Diffraction Tomography}\label{sec:bpm}

In optical diffraction tomography (ODT), the aim is to recover the refractive-index (RI) map of a sample from complex-valued measurements of the scattered fields generated when the sample is probed by a series of tilted incident fields \cite{wolf1969}. According to the scalar-diffraction theory, the propagation of the incident fields through the sample is governed by the wave equation. While pioneering works relied on linear models to approximate this propagation \cite{wolf1969,devaney1981inverse}, recent works have significantly improved the quality of RI reconstruction by using more accurate nonlinear models that account for multiple scattering \cite{soubies2017efficient}. Here, we look at one such nonlinear model called the beam-propagation method (BPM). \\ 

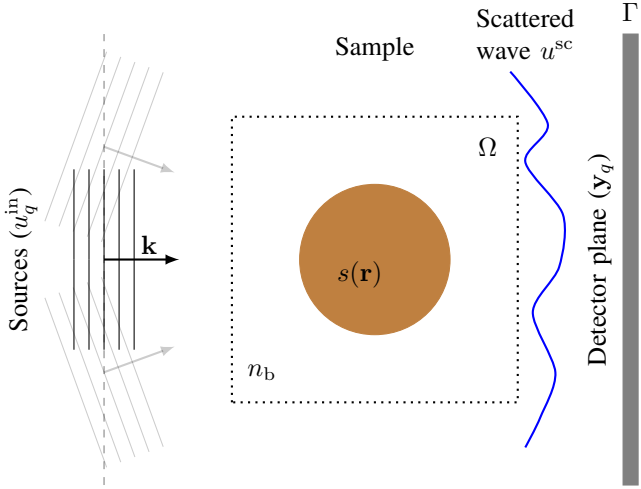
\begin{figure}[!t]
	\centering
	
	\begin{tikzpicture}
		\draw[dotted,thick]  (-1.9,-1.9) rectangle (1.9,1.9);
		\draw[gray,fill=gray]  (3.3,-3) rectangle (3.5,3);
		\draw[gray,dashed,thin]  (-3.6,-3) -- (-3.6,3);
		\filldraw[color=brown,fill=brown] (0,0) circle (1);
		  
		  \node at (-0.2,-0.2) {$s(\mathbf{r})$};
		  \node at (-1.5,-1.5) {$n_{\mathrm{b}}$};
		\draw[blue,thick] plot[smooth] coordinates {(1.8,2.5) (2.3,1.8) (2,1.3) (2.5,0.6)  (2.45,-0.1) (2.1,-0.7) (2.4,-1.5) (2.2,-2.1) (2,-2.5)};
		\node[align=center] at (2,3) {Scattered \\ wave $u^{\mathrm{sc}}$};
		\node[align=center] at (0,2.8) {Sample};
		\node[align=center] at (3.4,3.3) {$\Gamma$};
		\draw[-latex,thick,opacity=0.2] (-3.6,1.5)  -- ++ (-20:1cm);
		\draw[-latex,thick]  (-3.6,0) --  ++(0:1cm);
		\node at (-3,0.2) {$\mathbf{k}$};
		\draw[-latex,thick,opacity=0.2]  (-3.6,-1.5) -- ++(20:1cm);
		\draw  (-3.2,0) --  ++(90:1.2cm);\draw  (-3.2,0) --  ++(-90:1.2cm);
		\draw  (-3.4,0) --  ++(90:1.2cm);\draw  (-3.4,0) --  ++(-90:1.2cm);
		\draw  (-3.6,0) --  ++(90:1.2cm);\draw  (-3.6,0) --  ++(-90:1.2cm);
		\draw  (-3.8,0) --  ++(90:1.2cm);\draw  (-3.8,0) --  ++(-90:1.2cm);
		\draw  (-4,0) --  ++(90:1.2cm);\draw  (-4,0) --  ++(-90:1.2cm);
		\draw[opacity=0.2]   (-3.6,1.5)   ++ (160:0.4cm) --  ++(70:1.2cm);\draw[opacity=0.2]   (-3.6,1.5)   ++ (160:0.4cm)--  ++(-110:1.2cm);
		\draw[opacity=0.2]  (-3.6,1.5)   ++ (160:0.2cm)  --  ++(70:1.2cm);\draw[opacity=0.2]  (-3.6,1.5)   ++ (160:0.2cm) --  ++(-110:1.2cm);
		\draw[opacity=0.2]  (-3.6,1.5)   ++ (-20:0cm)  --  ++(70:1.2cm);\draw[opacity=0.2]  (-3.6,1.5)   ++ (-20:0cm) --  ++(-110:1.2cm);
		\draw[opacity=0.2]  (-3.6,1.5)   ++ (-20:0.2cm)  --  ++(70:1.2cm);\draw[opacity=0.2]   (-3.6,1.5)   ++ (-20:0.2cm)  --  ++(-110:1.2cm);
		\draw[opacity=0.2]  (-3.6,1.5)   ++ (-20:0.4cm)  --  ++(70:1.2cm);\draw[opacity=0.2]  (-3.6,1.5)   ++ (-20:0.4cm) --  ++(-110:1.2cm);
		\draw[opacity=0.2]   (-3.6,-1.5)   ++ (200:0.4cm)--  ++(110:1.2cm);\draw[opacity=0.2]   (-3.6,-1.5)   ++ (200:0.4cm) --  ++(-70:1.2cm);
		\draw[opacity=0.2]  (-3.6,-1.5)   ++ (200:0.2cm)--  ++(110:1.2cm);\draw[opacity=0.2]  (-3.6,-1.5)   ++ (200:0.2cm) --  ++(-70:1.2cm);
		\draw[opacity=0.2]  (-3.6,-1.5)   ++ (20:0cm) --  ++(110:1.2cm);\draw[opacity=0.2]  (-3.6,-1.5)   ++ (20:0cm) --  ++(-70:1.2cm);
		\draw[opacity=0.2]  (-3.6,-1.5)   ++ (20:0.2cm) --  ++(110:1.2cm);\draw[opacity=0.2]  (-3.6,-1.5)   ++ (20:0.2cm) --  ++(-70:1.2cm);
		\draw[opacity=0.2]  (-3.6,-1.5)   ++ (20:0.4cm) --  ++(110:1.2cm);\draw[opacity=0.2]  (-3.6,-1.5)   ++ (20:0.4cm) --  ++(-70:1.2cm);
		\node[rotate=90] at (3,0) {Detector plane ($\mathbf{y}_q$)};
		\node[rotate=90] at (-4.7,0) {Sources ($u_q^{\mathrm{in}}$)};
		\node at (1.5,1.5) {$\Omega$};
	\end{tikzpicture}
	\caption{\label{fig:schema} Optical diffraction tomography. A sample of refractive index $n_{\mathrm{b}} + s(\mathbf{r})$ is immersed in a medium of index $n_{\mathrm{b}}$ and illuminated by an incident plane wave (wave vector~$\mathbf{k}$). The interaction of the wave with the object produces scattered waves, which are recorded at the detector plane.}
\end{figure}

\noindent \textbf{Helmholtz Equation.} We consider a sample with a real-valued spatially varying refractive index that is immersed in a medium with constant refractive index $n_{\mathrm{b}}$, as shown in Figure \ref{fig:schema}. The RI distribution in the region of interest $\Omega = [0, L_{\mathrm{x}}] \times [0, L_{\mathrm{z}}]$ is represented as $n(\M r) = n_{\mathrm{b}} + s(\M r)$, where $\M r = (x, z)$ and $s(\M r)$ is the RI contrast. The sample is illuminated with an incident plane wave $u^{\text{in}}(\M r)$ of free-space wavelength $\lambda$, whose direction of propagation is specified by the wave vector $\V k$. The total field $u(\M r)$ that results from the interaction between the sample and the incident wave is then recorded at the positions $\{\M r_m\}_{m=1}^{M'}$ in the detector plane $\Gamma$ to yield the complex measurements $\M y \in \C^{M'}$. The interplay between the total field $u(\M r)$ at any point in space and the refractive index contrast $\delta n(\M r)$ is described by the Helmholtz equation
\begin{equation}
    \nabla^2 u(\M r) + k_0^2 n^2(\M r) u(\M r) = 0,
\end{equation}
where $k_0 = \frac{2 \pi}{\lambda}$. \\

\begin{figure}[t]
  \centering
  \includegraphics[width=0.45\textwidth]{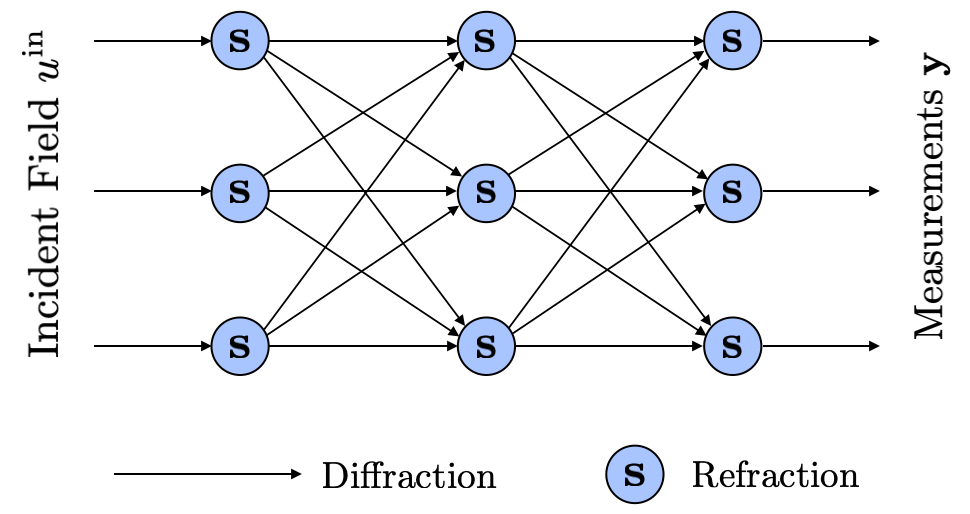}
 \caption{The computational structure for BPM resembles a neural network.}
 \label{fig:bpm_forward}
\end{figure} 

\noindent \textbf{Beam Propagation Method.} For computational purposes, the region of interest $\Omega$ is subdivided into an $(N_{\mathrm{x}} \times N_{\mathrm{z}})$ array of pixels with sampling steps $\delta_{\mathrm{x}}$ and $\delta_{\mathrm{z}}$ along the first and second dimension, respectively. The corresponding samples of the RI contrast $s(\M r)$ and total field $u(\M r)$ are stored in the vectors\footnote{Since the total field $u(\M r)$ depends on the RI contrast $s(\M r)$, we also refer to its discretized version as $\M u(\M s)$.} $\M s \in \R^K$ and $\M u \in \C^K$, respectively, where $K = N_{\mathrm{x}} N_{\mathrm{z}}$. Further, let $\M s_{k} \in \R^{N_{\mathrm{x}}}$ and $\M u_{k} \in \C^{N_{\mathrm{x}}}$ represent the above quantities when restricted to the slice $z=k \delta_{\mathrm{z}}$.

BPM computes the total field $\M u$ in a slice-by-slice manner along the \textit{z}-axis. For a given incident wave $u^{\text{in}}(\M r)$ that is propagated over a region larger than $\Omega$, we set the initial conditions as $\M u_{-1}(\M s) = \big(u^{\text{in}}(i \delta_{\mathrm{x}}, -\delta_{\mathrm{z}})\big)_{i=0}^{N_{\mathrm{x}}-1} \in \C^{N_{\mathrm{x}}}$. The total field over $\Omega$ is then computed via a series of diffraction and refraction steps
\begin{align}
    \widetilde{\M u}_{k}(\M s) &= \M u_{k-1}(\M s) * \M h_{\text{prop}}^{\delta_{\mathrm{z}}} \hspace{2.3cm} \text{(diffraction)}  \\
    \M u_{k}(\M s) &= \widetilde{\M u}_{k}(\M s) \odot \M p_{k}(\M s) \hspace{2.2cm} \ \ \text{(refraction)},
\end{align}
where $k=0, 1, \ldots, (N_{\mathrm{z}}-1)$, and the symbols $*$ and $\odot$ stand for convolution and pointwise multiplication, respectively. The convolution kernel $\M h_{\text{prop}}^{\delta_{\mathrm{z}}} \in \C^{N_{\mathrm{x}}}$ for the diffraction step is characterized in the Fourier domain as
\begin{equation}
    \V{\Spc F} \big\{ \M h_{\text{prop}}^{\delta_{\mathrm{z}}} \big \} (\M w_{\mathrm{x}}) = \mathrm{e}^{\mathrm{j} \delta_{\mathrm{z}} \Big(\sqrt{k_0^2 {n_{\mathrm{b}}}^2 \ - \ \M w_{\mathrm{x}}^2}\Big)},
\end{equation}
where $\V{\Spc F}$ denotes the discrete Fourier transform and $\M w_{\mathrm{x}} \in \R^{N_{\mathrm{x}}}$ is the frequency variable. The subsequent refraction step involves a pointwise multiplication with the phase mask
\begin{equation}
    \M p_{k}(\M s) = \mathrm{e}^{\mathrm{j} k_0 \delta_{\mathrm{z}} \M s_{k}}.
\end{equation}
Finally, we define an operator $\M R: \C^{N_{\mathrm{x}}} \mapsto \C^{M'}$ that propagates $\M u_{N_{\mathrm{z}}-1}(\M s)$ to the detector plane $\Gamma$ and restricts it to the sensor positions to give us the measurements $\M y \in \C^{M'}$. Thus, for a given incident wave $u^{\text{in}}$, our noise-free nonlinear BPM forward model is of the form 
\begin{equation}\label{eq:bpm_single_forward}
    \M y_0 = \M H_{\text{bpm}}(\M s; u^{\text{in}}) = \M R \big(\M u_{N_{\mathrm{z}}-1}(\M s) \big).
\end{equation}

In Figure \ref{fig:bpm_forward}, we show the implementation of $\M H_{\text{bpm}}$ as a directed acyclic graph. \\

\noindent \textbf{Complete Forward Model.} We assume that the sample is illuminated with $Q$ incident plane waves $\{u_{q}^{\text{in}}\}_{q \in \{1, \ldots, Q\}}$ and that the corresponding measurements are $\{\M y_q \in \C^{M'}\}_{q \in \{1, \ldots, Q\}}$. These measurements are related to the RI contrast $\M s$ of the sample through the BPM forward model in \eqref{eq:bpm_single_forward}. We define a stacked measurement vector as $\M y = (\M y_1, \ldots, \M y_Q) \in \R^M$ ($M = QM'$). This allows us to rewrite the complete forward model in the form of \eqref{eq:nonlinear_forward_general}, where the operator $\M H$ consists of the application of $\M H_{\text{bpm}}$ with all the illuminations and the concatenation of the outputs into a single vector.

\section{Bayesian Reconstruction Framework} \label{sec:bayesian_framework}
We now present our reconstruction framework that is based on Bayesian statistics for solving the generic nonlinear inverse problem described in Section \ref{subsec:forward_model_structure}. The image $\M s$ is assumed to be a realization of a random vector with pdf $p_{\mathrm{S}}$ and the statistical model for measurement noise is included within the likelihood function $p_{\mathrm{Y}|\mathrm{S}}$, which is the conditional distribution of the measurements given the image. The quantity of interest here is the posterior distribution $p_{\mathrm{S}|\mathrm{Y}}$ as it provides a complete statistical characterization of the problem at hand. Using Bayes' rule, we then write $p_{\mathrm{S}|\mathrm{Y}}$ as
\begin{equation}
  p_{\mathrm{S}|\mathrm{Y}}(\M s | \M y) = \frac{p_{\mathrm{Y}|\mathrm{S}}(\M y | \M s) p_{\mathrm{S}}(\M s)}{\int_{\R^K} p_{\mathrm{Y}|\mathrm{S}}(\M y | \M s) p_{\mathrm{S}}(\M s) \ud \M s}.
\end{equation}
In this section, we first characterize the likelihood function $p_{\mathrm{Y}|\mathrm{S}}$. We then discuss the prior distribution $p_{\mathrm{S}}$, which, in our framework, is defined through a deep generative model, followed by the posterior distribution $p_{\mathrm{S}|\mathrm{Y}}$. Finally, we detail a MCMC scheme to generate samples from the posterior distribution. This allows us to perform inference by computing point estimates and the uncertainties associated with them.

\subsection{Likelihood Function}
In our framework, we assume that the operator $\M N : \M y_0 \mapsto \M N(\M y_0)$ in \eqref{eq:nonlinear_forward_general} samples the noisy measurement vector $\M y$ from a conditional distribution $p_{\mathrm{Y}|\mathrm{Y_0}}$ according to 
\begin{equation}
  \M y \sim p_{\mathrm{Y}|\mathrm{Y_0}}\big(\cdot | \M y_0 = \M H(\M s)\big),
\end{equation} 
where $p_{\mathrm{Y}|\mathrm{Y_0}}$ models the statistics of the noise in the imaging system. Since our forward models $\M H$ are deterministic, the quantity $p_{\mathrm{Y}|\mathrm{S}}$ (a.k.a. the likelihood function) is given by
\begin{equation}
  p_{\mathrm{Y}|\mathrm{S}}(\M y | \M s) = p_{\mathrm{Y}|\mathrm{Y_0}}(\M y | \M y_0 = \M H(\M s)).
\end{equation}

In many imaging systems, there exist multiple independent sources of noise. It is therefore reasonable to assume an additive white-Gaussian-noise (AWGN) model, as dictated by the central limit theorem. There, the distribution $p_{\mathrm{Y}|\mathrm{Y_0}}$ is
\begin{equation}
  p_{\mathrm{Y}|\mathrm{Y_0}}(\M y | \M y_0) \propto \exp{\Bigg(-\frac{\|\M y - \M y_0\|_{2}^{2}}{2\sigma^2}\Bigg)},
\end{equation}
where $\sigma$ is the standard deviation of the Gaussian noise.

Another model that is commonly used is the shot- or Poisson-noise model. In this case, we have that
\begin{equation}
  p_{\mathrm{Y}|\mathrm{Y_0}}(\M y | \M y_0) = \prod_{m=1}^{M} \frac{([\M y_0]_m)^{[\M y]_m}}{([\M y]_m)!} \exp{\big(-[\M y_0]_m\big)},
\end{equation}
where $\M y \in \N^{M}$.

\subsection{Prior Distribution}
The choice of the distribution $p_{\mathrm{S}}$ reflects our prior knowledge about the image of interest. This information is crucial for the resolution of the inverse problem, especially when it is ill-posed. In classical Bayesian methods, $p_{\mathrm{S}}$ is chosen from a family of distributions with closed-form analytical expressions such that it fits the characteristics of the image and also allows for efficient inference. Popular examples include the Gaussian and Markovian models. In our framework, we instead propose to leverage the power of neural networks to define a data-driven prior distribution.

We assume that we have access to a dataset that contains sample images from the true (but unknown) probability distribution $p_{\text{image}}$ of our image of interest. The idea then is to approximate $p_{\text{image}}$ with $p_{\mathrm{S}}$ as defined by a deep generative model. More specifically, we consider generative models consisting of a generator network $\Op G: \R^d \rightarrow \R^K$ ($d \ll K$) that maps a low-dimensional latent space to the high-dimensional image space. This network takes a vector $\M z \in \R^d$, which is sampled from some distribution $p_{\mathrm{Z}}$ (typically a Gaussian or uniform distribution), and outputs a sample image $\Op G(\M z)$. Thus, the generator network $\Op G$ and the distribution $p_{\mathrm{Z}}$ implicitly characterize $p_{\mathrm{S}}$ and provide us with a way to directly sample from it. If this model is properly trained, the resulting $p_{\mathrm{S}}$ is close to $p_{\text{image}}$ and the generated images are statistically similar to the ones in the dataset. 

In our experiments (see Section \ref{sec:experiments}), we use the well-known Wasserstein GANs (WGANs) \cite{arjovsky2017wasserstein} for our data-driven prior. We provide a brief description of WGANs in Appendix \ref{app:GANs}. \\ 

\noindent \textbf{Augmented Deep Generative Priors.} 
The training of deep generative models such as GANs requires large amounts of data and is a challenging task in general. Over the past few years, there have been several proposals for performance improvements that have led to the development of better training schemes and network architectures. Most existing works use normalized datasets, where each image has the same range of pixel values. However, this is not suitable if we wish to use such models as priors in quantitative imaging (\textit{e.g.,} ODT). In these modalities, it is important to recover the actual values of the object (image) as compared to only the contrast. Thus, we require our generative model to be able to output images with different ranges of pixel values.   

While performing our experiments, we observed that the training of high-quality WGANs on unnormalized datasets was non-trivial. We propose a simple effective workaround, which simplifies the training and allows us to build models that generate images with different ranges. We define an augmented generative model $\Op G_{h}: \R^{d+1} \rightarrow \R^K$ ($d \ll K$) that consists of a (standard) generative network $\Op G: \R^{d} \rightarrow \R^K$ trained on a normalized dataset and a deterministic function $h: \R \rightarrow \R$. Here, the latent vector $\M z = (\M z_1, z_2) \in \R^{d+1}$ has two independent components $\M z_1 \in \R^{d}$ and $z_2 \in \R$ that are sampled from $p_{\mathrm{Z_1}}$ and $p_{\mathrm{Z_2}}$, respectively. The output image is given by $\Op G_{h}(\M z) = h(z_2) \Op G (\M z_1)$. For a generated image $\Op G_{h}(\M z) \in \R^K$, the term $\Op G (\M z_1) \in \R^K$ represents its details or contrast, and the term $h(z_2)$ represents its scaling factor. Since $\Op G$ is now required to only produce images with the same range, we can rely on existing GANs to obtain high-quality models. Moreover, the distribution of the scaling factor can be easily controlled by carefully choosing the distribution $p_{\mathrm{Z_2}}$ and the function $h$.

\subsection{Posterior Distribution}\label{sec:post}
Now that we are equipped with the likelihood function $p_{\mathrm{Y}|\mathrm{S}}$ and the prior distribution $p_{\mathrm{S}}$, we look at the posterior distribution $p_{\mathrm{S}|\mathrm{Y}}$ of the image. Since our prior distribution $p_{\mathrm{S}}$ is defined by a pre-trained augmented deep generative model $\Op G_{h}: \R^{d+1} \rightarrow \R^K , \M z \mapsto \Op G_{h}(\M z)$ with $p_{\mathrm{Z}}(\M z) = p_{\mathrm{Z_1}}(\M z_1) p_{\mathrm{Z_2}}(z_2)$ for any $\M z = (\M z_1, z_2) \in \R^{d+1}$, our $p_{\mathrm{S}|\mathrm{Y}}$ is given by the push-forward of the posterior distribution $p_{\mathrm{Z}|\mathrm{Y}}$ of the latent vector through the mapping $\Op G_{h}$. The distribution $p_{\mathrm{Z}|\mathrm{Y}}$ can be written as
\begin{align*}
    p_{\mathrm{Z}|\mathrm{Y}}(\M z | \M y) &= \frac{p_{\mathrm{Y}|\mathrm{Z}}(\M y | \M z) p_{\mathrm{Z}}(\M z)}{\int_{\R^{d+1}} p_{\mathrm{Y}|\mathrm{Z}}(\M y | \M z) p_{\mathrm{Z}}(\M z) \ud \M z},
\end{align*}
where $p_{\mathrm{Y} | \mathrm{Z}}(\M y | \M z) = p_{\mathrm{Y}|\mathrm{S}}(\M y | \M s = \Op G_h(\M z))$.

A Bayesian inverse problem is said to be well-posed in some metric on the space of probability measures if its solution (the posterior distribution) exists, is unique, and is continuous with respect to the measurements for the chosen metric \cite{latz2020well}. Depending on the metric, the well-posedness of the Bayesian inverse problem ensures continuity of posterior expectations of appropriate quantities of interest. Based on the work in \cite{latz2020well}, we can show that for the AWGN model, our Bayesian problem is well-posed in the Prokhorov, total-variation and Hellinger distances. Moreover, our problem is well-posed in the Wasserstein distance if $p_{\mathrm{Z}}$ satisfies a finite-moment-like condition. By using a result from \cite{holden2022bayesian}, we can also show the existence of the moments of our posterior distribution under mild conditions on $p_{\mathrm{Z}}$ and $\Op G_h$. We provide the details regarding these properties in Appendix \ref{app:prop_post}.

\subsection{Sampling from the Posterior Distribution}\label{sec:mala}
The proposed framework allows one to draw samples in the low-dimensional latent space instead of the high-dimensional image space directly. Specifically, if we generate a sample $\overline{\M z}$ from $p_{\mathrm{Z}|\mathrm{Y}}$, then the image $\overline{\M s} = \Op G_{h}(\overline{\M z})$ is a sample from $p_{\mathrm{S}|\mathrm{Y}}$.

In this work, we use the Metropolis-adjusted Langevin algorithm (MALA) \cite{roberts1996exponential,roberts2002langevin}, which is a MCMC method, to sample from $p_{\mathrm{Z}|\mathrm{Y}}$. Given a sample $\overline{\M z}_{t}$, MALA generates $\overline{\M z}_{t+1}$ in two steps. In the first step, we construct a proposal $\widetilde{\M z}_{t+1}$ for the new sample according to
\begin{equation}
 \widetilde{\M z}_{t+1} = \overline{\M z}_{t} + \eta \nabla_{\M z} \log p_{\mathrm{Z}|\mathrm{Y}}(\overline{\M z}_{t} | \M y) + \sqrt{2 \eta} \boldsymbol{\zeta},
\end{equation}
where $\boldsymbol{\zeta}$ is drawn from the standard multivariate Gaussian distribution and $\eta \in \R_{+}$ is a fixed step-size. In the second step, the proposal $\widetilde{\M z}_{t+1}$ is either accepted or rejected, the acceptance probability being
\begin{equation}
  \alpha = \min \Big\{1, \frac{p_{\mathrm{Z}|\mathrm{Y}}(\widetilde{\M z}_{t+1} | \M y) q_{\M y}(\overline{\M z}_{t} | \widetilde{\M z}_{t+1})} {p_{\mathrm{Z}|\mathrm{Y}}(\overline{\M z}_{t} | \M y) q_{\M y}(\widetilde{\M z}_{t+1} | \overline{\M z}_{t})}  \Big\},
\end{equation}
where $q_{\M y}(\overline{\M z} | \widetilde{\M z}) = \exp\big(-\frac{1}{4 \eta} \|\overline{\M z} - \widetilde{\M z} - \eta \nabla_{\M z} \log p_{\mathrm{Z}|\mathrm{Y}}(\widetilde{\M z} | \M y) \|_{2}^{2}\big)$. If the proposal is accepted, then we set $\overline{\M z}_{t+1} = \widetilde{\M z}_{t+1}$; otherwise, $\overline{\M z}_{t+1} = \overline{\M z}_{t}$. One advantage of MALA is that it uses the gradient of the (log) target distribution to construct more probable proposals. In doing so, it explores the target distribution faster than some other MCMC methods such as the well-known random walk Metropolis-Hastings algorithm \cite{gelman1997weak}.

The major computational bottleneck in MALA is the computation of the gradient term $\nabla_{\M z} \log p_{\mathrm{Z}|\mathrm{Y}}$ as it involves terms such as ${\M J^{H}_{\M H}(\M x_1)} \M r_1$ and ${\M J^{H}_{\Op G_h}(\M x_2)} \M r_2$, where $\M x_1 \in \R^K$, $\M r_1 \in \C^M$, $\M x_2 \in \R^{d+1}$, and $\M r_2 \in \R^K$. For instance, if we assume an AWGN model with variance $\sigma^2$ and that $p_{\mathrm{Z}}$ is the standard mutivariate Gaussian distribution, then $p_{\mathrm{Z}|\mathrm{Y}}$ can be written as
\begin{equation}
  p_{\mathrm{Z}|\mathrm{Y}}(\M z | \M y) = \frac{1}{C} \exp\bigg(-\frac{\|\M y - \M H \{\Op G_h(\M z)\} \|_{2}^{2}}{2 \sigma^2} - \frac{\|\M z\|_{2}^{2}}{2}\bigg),
\end{equation}
where $C$ is the normalization factor. In this case, the gradient term is
\begin{equation}
  \nabla_{\M z} \log p_{\mathrm{Z}|\mathrm{Y}}(\M z | \M y) = -\frac{\M J^{H}_{\Op G_{h}}(\M z) \M J^{H}_{\M H}(\Op G_{h}(\M z)) (\M y - \M H\{\Op G_{h}(\M z)\})}{\sigma^2} - \M z.
\end{equation}
Since $\Op G_{h}$ is a neural network and $\M H$ has a neural-network-like structure, we then compute $\nabla_{\M z} \log p_{\mathrm{Z}|\mathrm{Y}}$ efficiently using an error backpropagation algorithm.

Once we have obtained the samples $\{\overline{\M z}_{t}\}_{t=1}^{T}$ from $p_{\mathrm{Z}|\mathrm{Y}}$, we transform them to get the samples $\{ \Op G_h(\overline{\M z}_{t}) \}_{t=1}^{T}$ from $p_{\mathrm{S}|\mathrm{Y}}$ and use them to perform inference. Specifically, we approximate any integral of the form $\int_{\R^K} f(\M s) p_{\mathrm{S}|\mathrm{Y}}(\M s | \M y) \ud \M s$, where $f:\R^K \rightarrow \R$ is a real-valued function, by its empirical estimate $E_{T}(f) = \frac{1}{T} \sum_{t=1}^{T}f(\Op G_h(\overline{\M z}_{t}))$. 

In practice, we discard some of the samples generated at the beginning of the chain to correct for their bias. This ``burn-in'' period can often be shortened by choosing a suitable starting point for the chain. We propose to initialize MALA with 
\begin{equation}
  \M z_{\text{init}} = \argmin_{\M z \in \R^{d+1}} \|\M s_{\text{init}} - \Op G_h(\M z) \|_{2}^{2},
\end{equation}
where $\M s_{\text{init}}$ is a low-quality estimate obtained by using some fast classical reconstruction algorithm. 

\section{Results and Discussion} \label{sec:experiments}
In this section, we show the benefits of our neural-network-based Bayesian reconstruction framework by applying it to both phase retrieval and optical diffraction tomography.

\subsection{Augmented WGANs}\label{sec:exp_aug_gans}

\begin{figure*}[t]
  \subcaptionbox{WGAN \vspace{0.5cm}}{\includegraphics[trim={0 0 0 0},width=\textwidth]{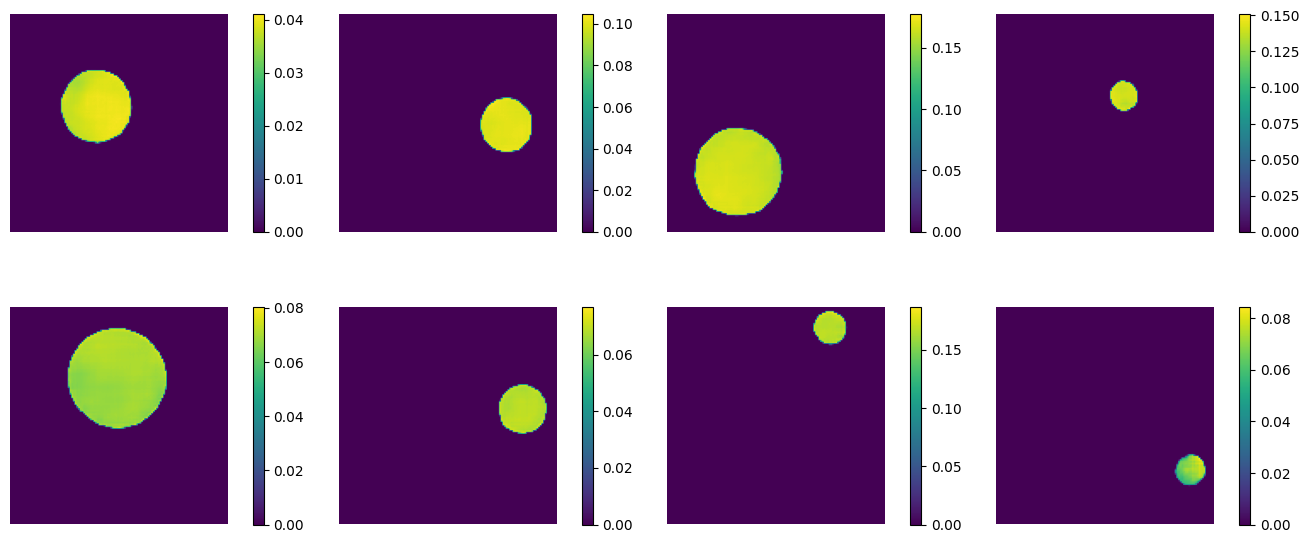}}
  \subcaptionbox{Augmented WGAN \vspace{0.5cm}}{\includegraphics[trim={0 0 0 0},width=\textwidth]{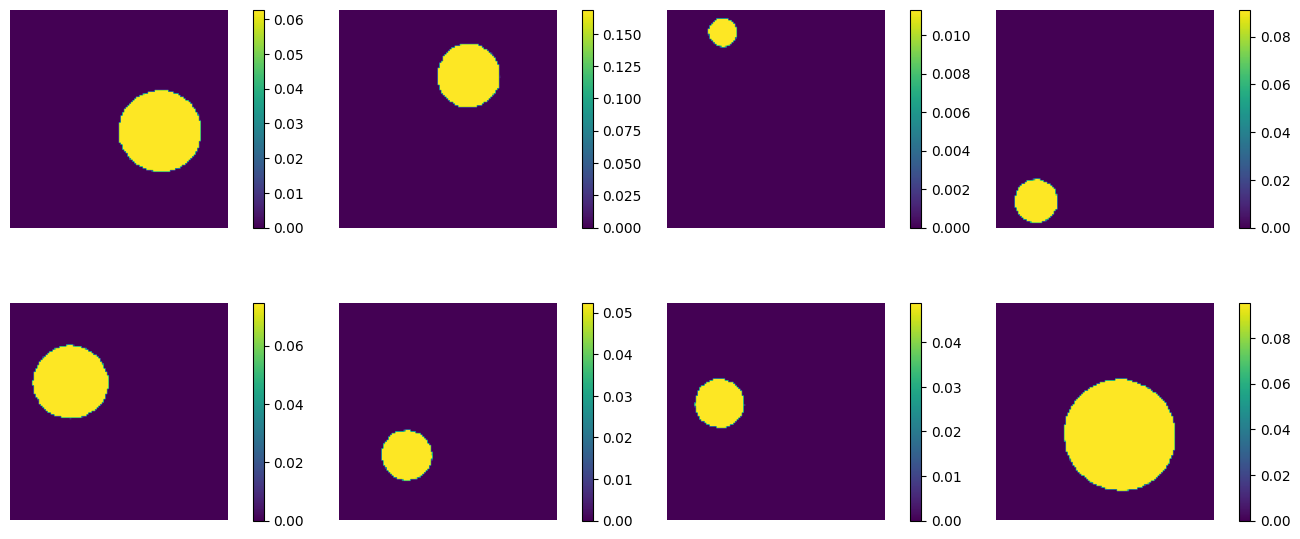}}
  \caption{Samples generated by trained models.}
  \label{fig:wgan_models}
\end{figure*}

In our first experiment, we highlight the importance of the proposed augmented generative models. We consider the task of training WGAN models on synthetic datasets consisting of $(128 \times 128)$ images, where each image contains a constant-valued disc and its background pixels are zero-valued. The coordinates $(x, y)$ of the center of the disc, its radius $r$ (in pixels), and its constant-intensity value $v$ follow the uniform distributions $U_{(10,115)}$, $U_{(10,115)}$, $U_{[8,35]}$, and $U_{(0,0.2]}$, respectively. The aforementioned parameters implicitly define the probability distribution $p_{\text{data}}$ that we wish to approximate using WGANs.

We qualitatively compare the performance of two models. The first model is a WGAN trained on $50,\!000$ images sampled from $p_{\text{data}}$. In this case, the distribution $p_{\mathrm{Z}}$ for the latent variable is chosen to be the standard multivariate Gaussian distribution. The second model is an augmented WGAN, where the WGAN component is trained on a normalized dataset with $50,\!000$ images. Thus, we first sample $50,\!000$ images from $p_{\text{data}}$ and we then normalize each of them such that the value of the disc is one. The distributions $p_{\mathrm{Z_1}}$ and $p_{\mathrm{Z_2}}$ are chosen to be standard Gaussian distributions as well, and the function $h$ is
\begin{equation}\label{eq:scaling_function}
  h(x) = \frac{0.2}{\sqrt{2 \pi}} \int_{-\infty}^{x} \mathrm{e}^{-\frac{t^2}{2}} \ud t. 
\end{equation} 
This choice of $h$ and $p_{\mathrm{Z_2}}$ ensures that the scaling factor of the augmented WGAN follows the uniform distribution $U_{(0,0.2]}$. For both the models, we use the generator and critic network architectures described in Appendix \ref{app:architectures}. The WGAN is trained for $2500$ epochs while the augmented WGAN is trained for $1250$ epochs using RMSProp optimizers with a learning rate of $5 \times 10^{-5}$ and a batch size of $64$. The parameters $\lambda_{\text{gp}}$ and $n_{\text{critic}}$ (refer to Appendix \ref{app:GANs}) are set as $10$ and $5$, respectively. 

In Figure \ref{fig:wgan_models}, we present typical samples generated by the two models. We observe that the augmented WGAN, unlike the WGAN, is able to produce sharp constant-valued discs.

\subsection{Phase Retrieval}
\begin{figure*}[t]
  \centering
  \begin{tikzpicture}[xscale=5,yscale=2.2]
    \node (A) at (0,0) {\includegraphics[scale=0.45]{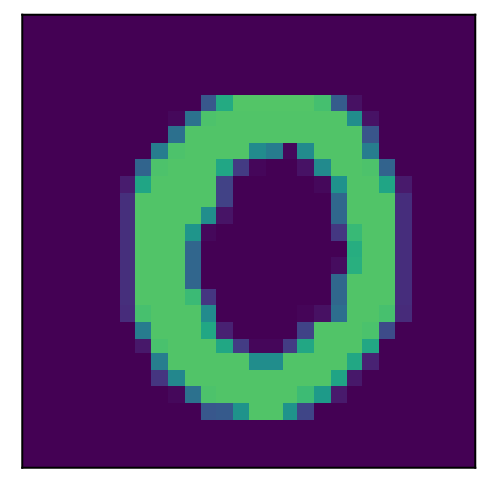}};
    \node at (A.north) {\small Ground-truth image};
    \node (B) at (1,1) {\includegraphics[scale=0.45]{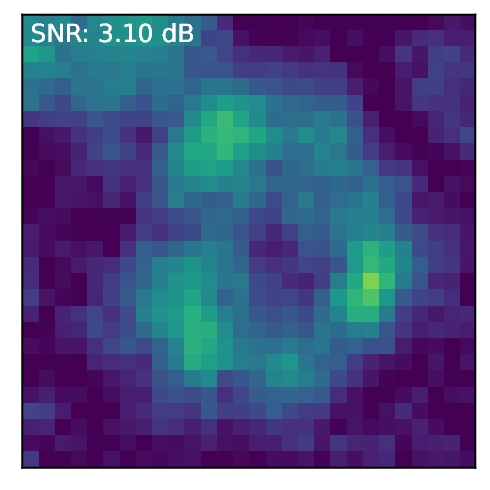}};
    \node at (B.north) {\small Initial reconstruction};
    \node (C) at (2,1) {\includegraphics[scale=0.45]{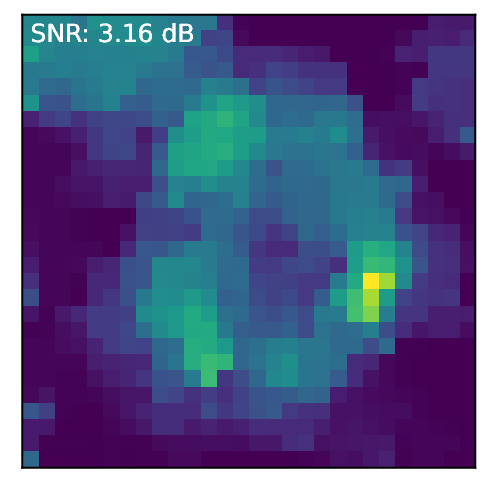}};
    \node at (C.north) {\small TV reconstruction};
    \node at (2.75,1) {\includegraphics[scale=0.45]{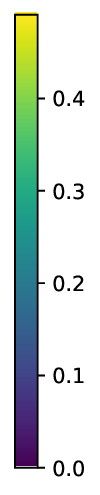}};
    \node (D) at (1,-1) {\includegraphics[scale=0.45]{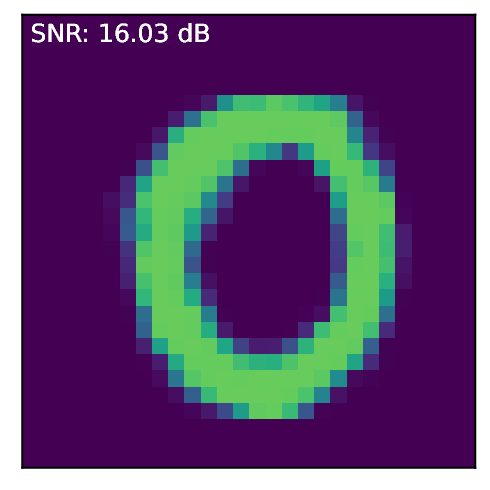}};
    \node at (D.north) {\small Posterior mean};
    \node (E) at (2,-1) {\includegraphics[scale=0.45]{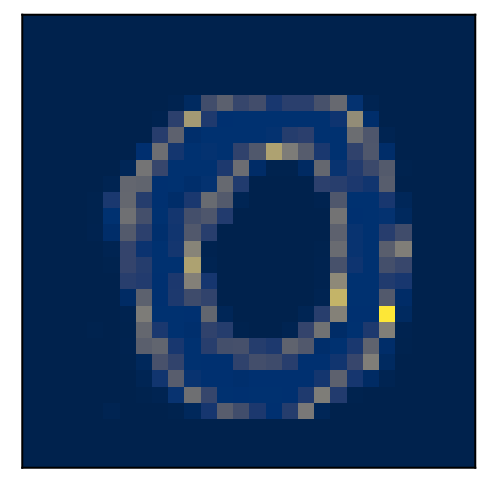}};
    \node at (E.north) {\small Posterior standard deviation};
    \node at (2.77,-1) {\includegraphics[scale=0.45]{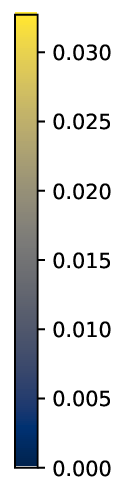}};
  \end{tikzpicture}
  \caption{Reconstructions for phase retrieval (oversampling ratio $M/K = 0.1$).\\}
  \label{fig:PR_res_1}
\end{figure*}
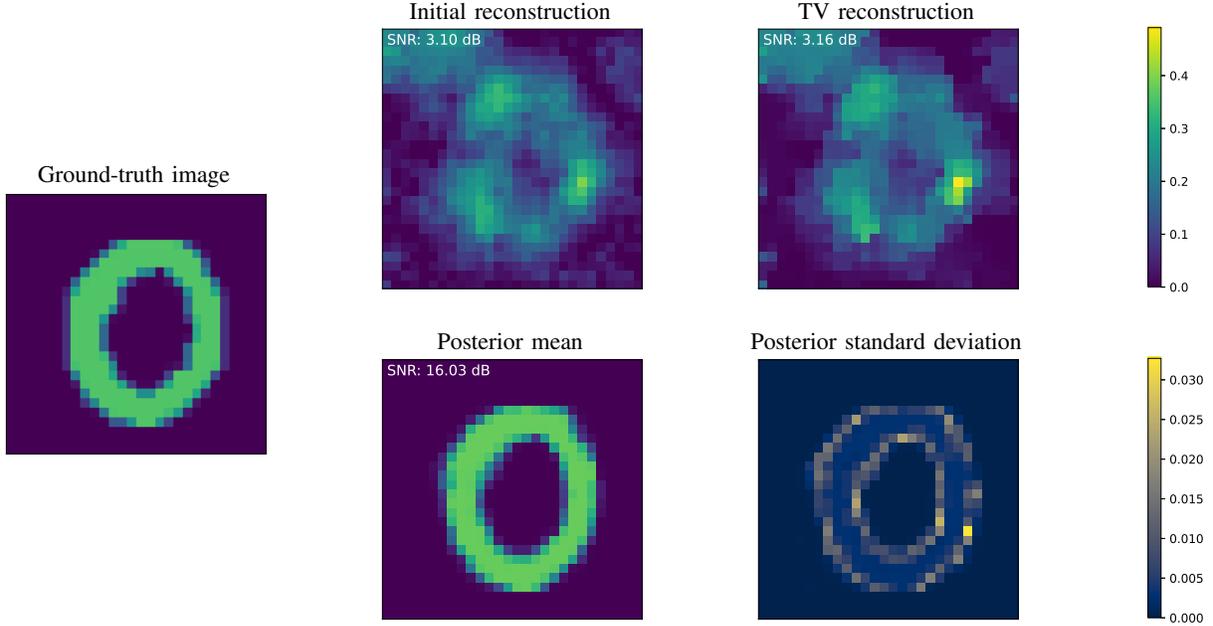

Next, we look at the phase-retrieval problem. We present two examples where the ground-truth images are taken from the MNIST \cite{lecun1998gradient} and Fashion-MNIST \cite{xiao2017/online} testing datasets. In both cases, the measurements $\M y \in \N^M$ are simulated according to \eqref{eq:phase_retrieval_forward} with a Poisson-noise model, where $\M A$ is one realization of a random matrix with i.i.d. entries from a zero-mean Gaussian distribution with variance $\sigma_{\M A}^2$.

\subsubsection{MNIST} The MNIST dataset contains $(28 \times 28)$ images of handwritten digits. The ground-truth image (Figure \ref{fig:PR_res_1}) is first normalized to have values in the range $[0,1]$ and is then multiplied by a factor $\alpha$ which is picked uniformly at random from $(0, 0.5]$.

In this case, the WGAN component of our augmented model $\Op G_h$ is trained on the normalized MNIST training dataset which contains $50,\!000$ images with values in the range $[0,1]$. The distributions $p_{\mathrm{Z_1}}$ and $p_{\mathrm{Z_2}}$ are standard Gaussian distributions and the function $h$ is
\begin{equation}\label{eq:scaling_function_pr}
    h(x) = \frac{0.5}{\sqrt{2 \pi}} \int_{-\infty}^{x} \mathrm{e}^{-\frac{t^2}{2}} \ud t. 
  \end{equation}
The architectures for the generator and critic networks can be found in Appendix \ref{app:architectures}. The WGAN is trained for $2000$ epochs using ADAM optimizers \cite{adam2014} with a learning rate of $2 \times 10^{-4}$, hyperparameters $(\beta_1, \beta_2) = (0.5, 0.999)$, and a batch size of $64$. The parameters $\lambda_{\text{gp}}$ and $n_{\text{critic}}$ are set as $10$ and $5$, respectively.

\subsubsection{Fashion-MNIST} 
\begin{figure*}[t]
  \centering
  \begin{tikzpicture}[xscale=5,yscale=2.2]
    \node (A) at (0,0) {\includegraphics[scale=0.45]{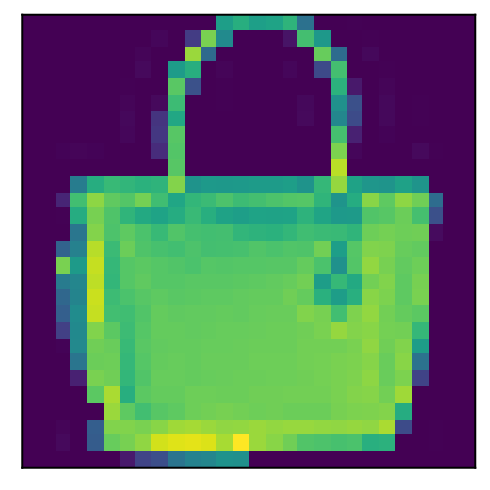}};
    \node at (A.north) {\small Ground-truth image};
    \node (B) at (1,1) {\includegraphics[scale=0.45]{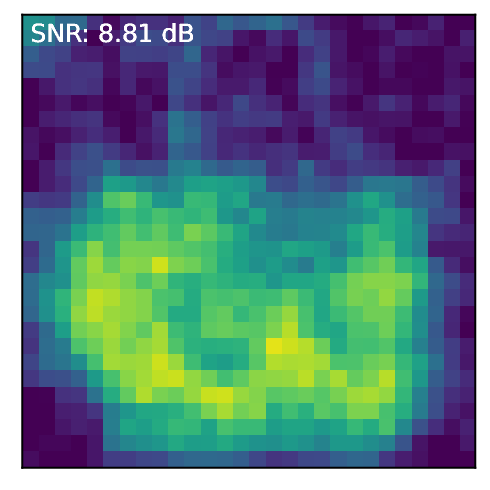}};
    \node at (B.north) {\small Initial reconstruction};
    \node (C) at (2,1) {\includegraphics[scale=0.45]{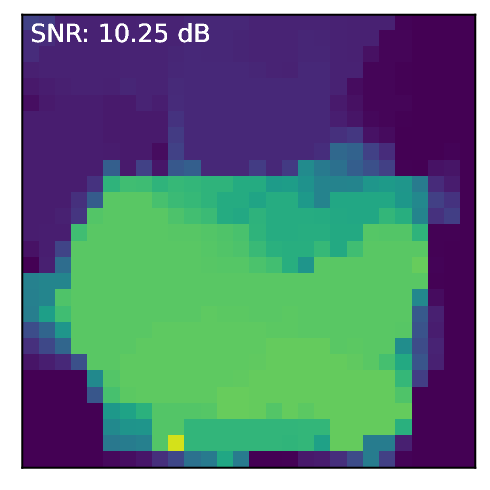}};
    \node at (C.north) {\small TV reconstruction};
    \node at (2.75,1) {\includegraphics[scale=0.45]{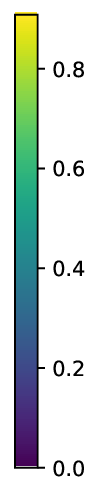}};
    \node (D) at (1,-1) {\includegraphics[scale=0.45]{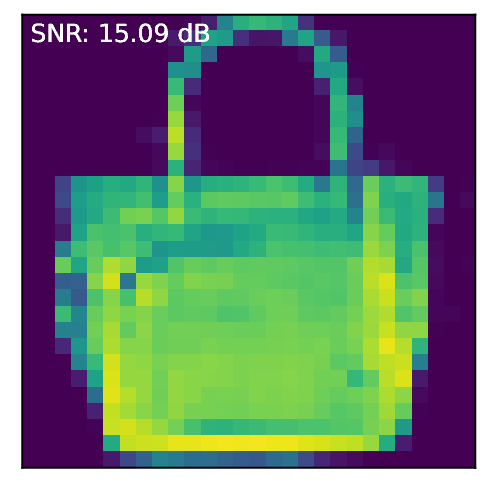}};
    \node at (D.north) {\small Posterior mean};
    \node (E) at (2,-1) {\includegraphics[scale=0.45]{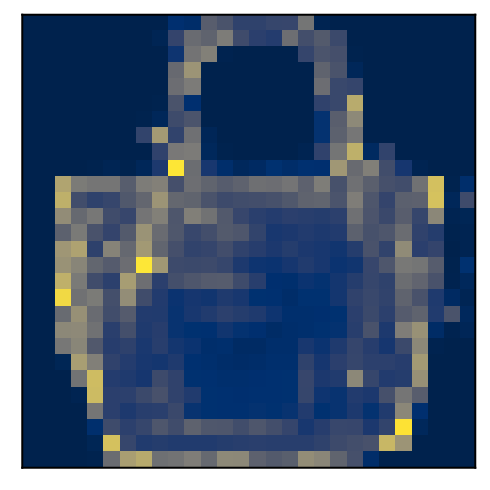}};
    \node at (E.north) {\small Posterior standard deviation};
    \node at (2.76,-1) {\includegraphics[scale=0.45]{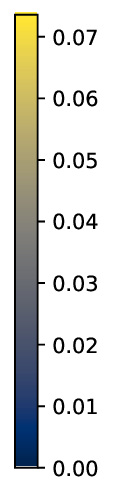}};
  \end{tikzpicture}
  \caption{Reconstructions for phase retrieval (oversampling ratio $M/K = 0.15$).\\}
  \label{fig:PR_res_2}
\end{figure*}
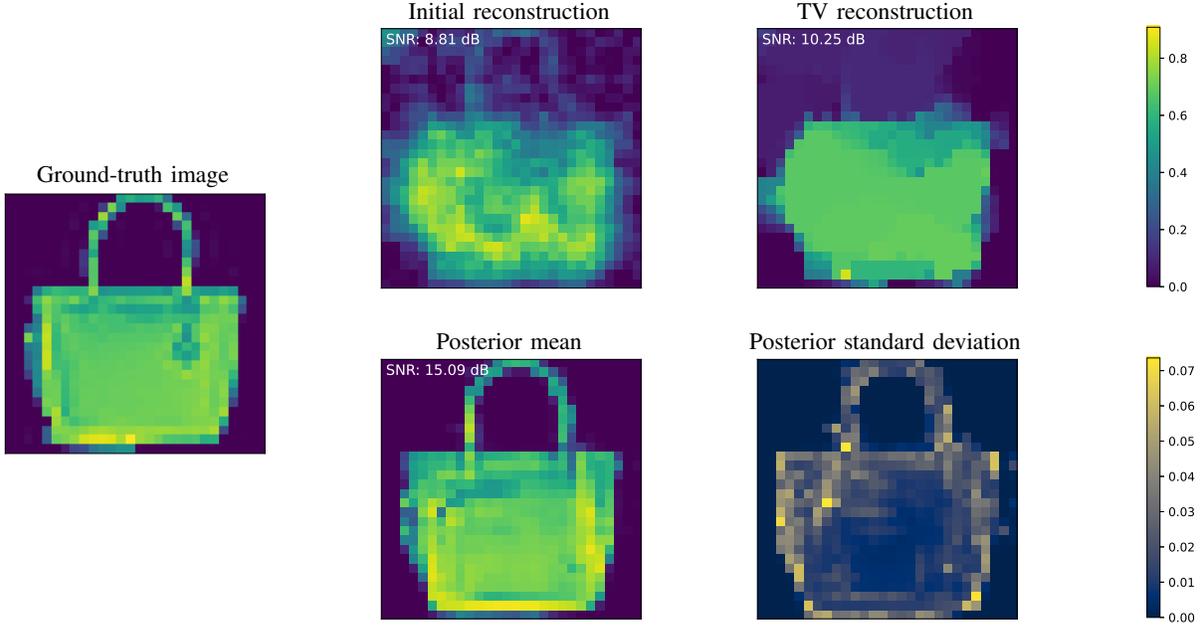

The Fashion-MNIST dataset consists of $(28 \times 28)$ grayscale images of different fashion products. Our ground-truth image from this dataset is shown in Figure \ref{fig:PR_res_2}.

Here, the WGAN for our augmented deep generative prior is trained on the normalized Fashion-MNIST training dataset. It contains $60,\!000$ images whose values lie in the range $[0,1]$. The distributions $p_{\mathrm{Z_1}}$ and $p_{\mathrm{Z_2}}$ are taken as standard Gaussian distributions while the function $h$ is
\begin{equation}\label{eq:scaling_function_pr}
    h(x) = \frac{1}{\sqrt{2 \pi}} \int_{-\infty}^{x} \mathrm{e}^{-\frac{t^2}{2}} \ud t. 
\end{equation}
We provide the architectures for the generator and critic networks in Appendix \ref{app:architectures}. The WGAN is trained for $2250$ epochs using ADAM optimizers with a learning rate of $2 \times 10^{-4}$, hyperparameters $(\beta_1, \beta_2) = (0.5, 0.999)$, and a batch size of $64$. The parameters $\lambda_{\text{gp}}$ and $n_{\text{critic}}$ are set as $10$ and $5$, respectively.

\subsubsection{Methods} As discussed in Section \ref{sec:mala}, we draw samples from the posterior distribution using MALA. The estimate $\M s_{\text{init}}$ that we use for initializing the chain is taken to be the solution of a variational problem with Tikhonov regularization, as in
\begin{align}\label{eq:pr_tikhonov}
  \M s_{\text{init}} = &\argmin_{\M s \in \R^K} \Bigg( \sum_{m=1}^{M} \bigg( -[\M y]_m \log \Big( \big[|\M A \M s|^{2} \big]_m \Big) + \big[|\M A \M s|^{2}\big]_m  \bigg) \nonumber \\ 
  & \mbox{ } + \ \tau \|\nabla \M s \|_{2,2}^{2} \ + \ i_{+}(\M s) \Bigg).
\end{align} 
There, $\nabla: \R^K \rightarrow \R^{K \times 2}$ is the gradient operator, $\|\cdot\|_{p,q}$ is the ($\ell_p, \ell_q$)-mixed norm defined as
\begin{equation}
  \| \M x \|_{p,q} \eqdef \bigg (\sum_{u=1}^{U} \bigg(\sum_{v=1}^{V} \big([\M x]_{u, v}\big)^{p} \bigg)^{q/p} \bigg)^{1/q} \ \ \forall \M x \in \R^{U \times V},
\end{equation}
$\tau \in \R_{+}$ is the regularization parameter and the functional $i_{+}$ given by
\begin{equation}
  i_{+}(\M s) = \begin{cases}
    0, \ &\M s \in \R_{+}^{K}\\
    + \infty, \ &\text{otherwise}
  \end{cases}
\end{equation} 
enforces the non-negativity constraint on the solution. The data-fidelity term in \eqref{eq:pr_tikhonov} corresponds to the negative log-likelihood under the Poisson-noise model. We solve the problem in \eqref{eq:pr_tikhonov} using a projected-gradient-descent algorithm. The regularization parameter $\tau$ so that it minimizes the mean-square error (MSE) with respect to the ground-truth is chosen via grid search.

After discarding the first $T_{\mathrm{b}}$ samples (burn-in period), we collect the next $T$ samples for performing inference. We compute the posterior mean which corresponds to the minimum mean-square error (MMSE) estimate. Further, to quantify the uncertainty associated with our estimation, we also compute the pixel-wise standard-deviation map. 

We compare the performance of our GAN-based posterior-mean estimator with that of the TV-regularized method \cite{rudin1992nonlinear}
\begin{align}\label{eq:pr_tv}
  \M s_{\text{TV}} = &\argmin_{\M s \in \R^K} \Bigg( \sum_{m=1}^{M} \bigg( -[\M y]_m \log \Big( \big[|\M A \M s|^{2} \big]_m \Big) + \big[|\M A \M s|^{2}\big]_m  \bigg) \nonumber \\ 
  & \mbox{ } + \ \tau \|\nabla \M s \|_{2,1} \ + \ i_{+}(\M s) \Bigg).
\end{align} 
TV regularization is known to promote piecewise-constant solutions and is well-matched to our test images. We solve \eqref{eq:pr_tv} using FISTA \cite{beck2009fast} initialized with $\M s_{\text{init}}$. The regularization parameter $\tau$ is tuned for optimal MSE performance with the help of a grid search.

\subsubsection{Results} To illustrate the advantage of our neural-network-based prior, we consider extreme imaging settings where the number of measurements $M$ is very small. For the first case (Figure \ref{fig:PR_res_1}), we have that $\alpha = 0.36, M/K = 0.1, \sigma_{\M A}^{2} = 10, \eta = 10^{-5}, T_{\mathrm{b}} = 8 \times 10^{5}$, and $T = 12 \times 10^{5}$. The parameters for the second case (Figure \ref{fig:PR_res_2}) are $M/K = 0.15, \sigma_{\M A}^{2} = 0.5, \eta = 1.75 \times 10^{-6} , T_{\mathrm{b}} = 17.5 \times 10^{5}$, and $T = 5 \times 10^{5}$.

In Figures \ref{fig:PR_res_1} and \ref{fig:PR_res_2}, we see that the GAN-based posterior-mean estimator outperforms the TV-regularized method considerably. Here, the very low oversampling ratios severely affect the performance of TV regularization, even though it is a good fit for the underlying images. By contrast, despite the scarcity of measurements, our estimator remarkably yields excellent results. This highlights the potential of learning-based priors for highly ill-posed problems. Finally, we observe that, as one would expect, the standard-deviation maps indicate higher uncertainty at the edges for the posterior-mean estimator.

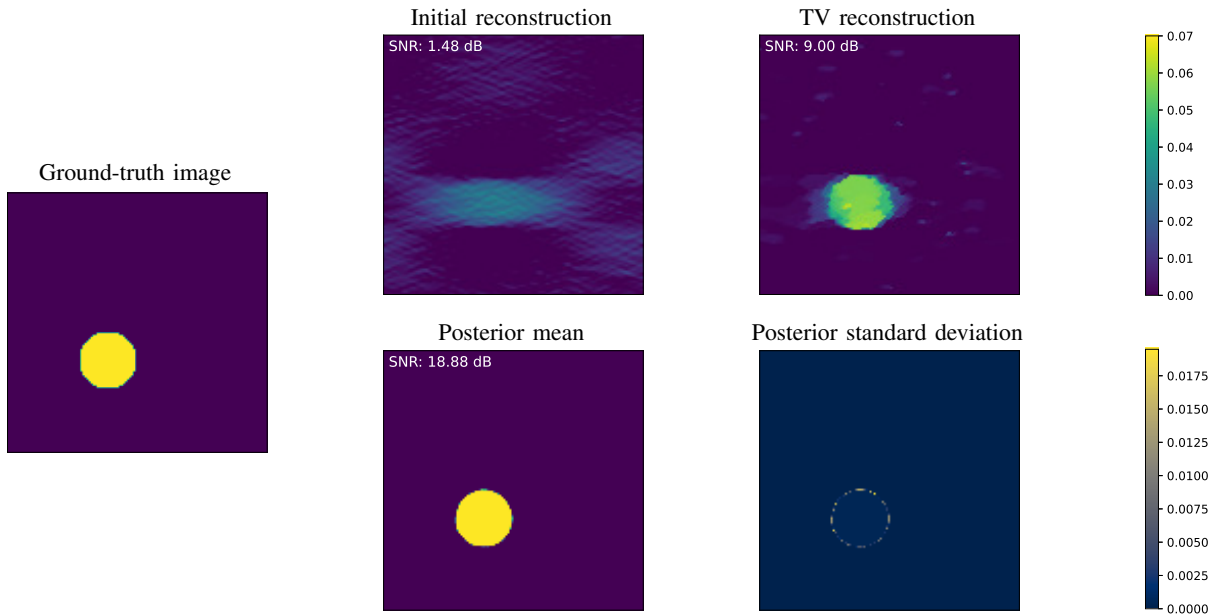
\begin{figure*}[t]
  \centering
  \begin{tikzpicture}[xscale=5,yscale=2.1]
    \node (A) at (0,0) {\includegraphics[scale=0.45]{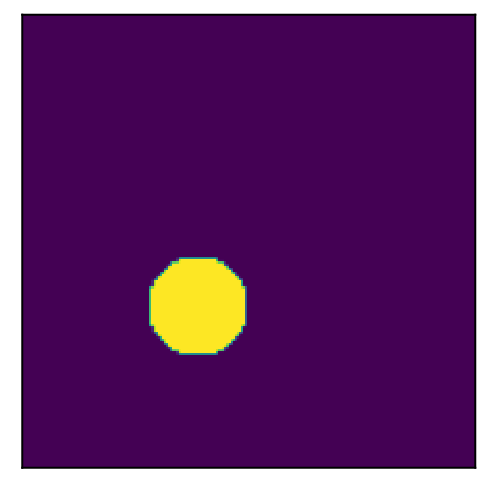}};
    \node at (A.north) {\small Ground-truth image};
    \node (B) at (1,1) {\includegraphics[scale=0.45]{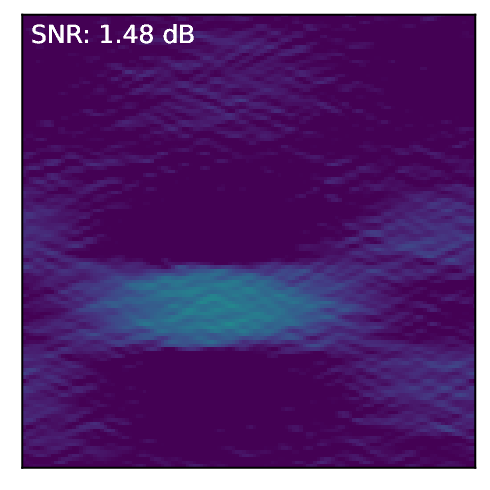}};
    \node at (B.north) {\small Initial reconstruction};
    \node (C) at (2,1) {\includegraphics[scale=0.45]{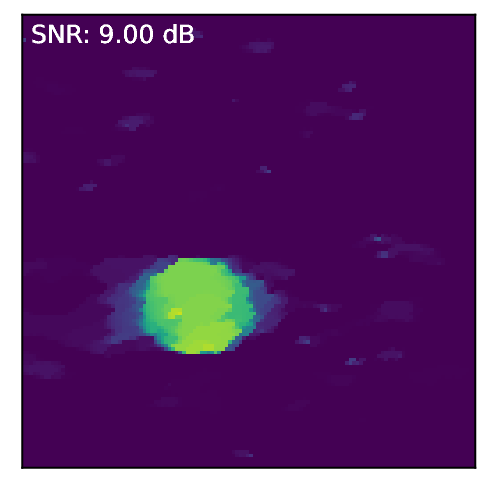}};
    \node at (C.north) {\small TV reconstruction};
    \node at (2.75,1) {\includegraphics[scale=0.45]{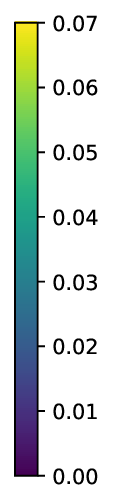}};
    \node (D) at (1,-1) {\includegraphics[scale=0.45]{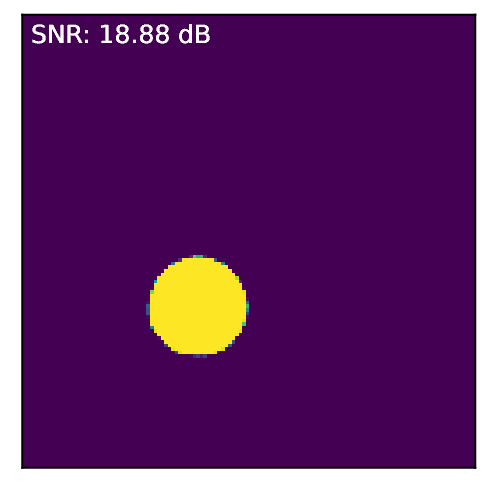}};
    \node at (D.north) {\small Posterior mean};
    \node (E) at (2,-1) {\includegraphics[scale=0.45]{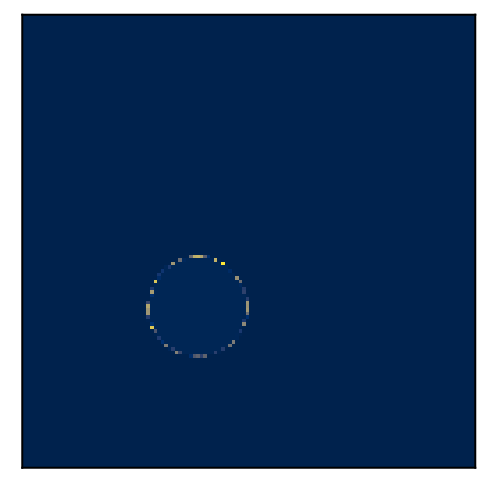}};
    \node at (E.north) {\small Posterior standard deviation};
    \node at (2.77,-1) {\includegraphics[scale=0.45]{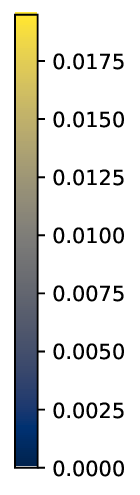}};
  \end{tikzpicture}
  \caption{Reconstructions for ODT ($v=0.07$).}
  \label{fig:ODT_res_1}
\end{figure*}

\subsection{Optical Diffraction Tomography}
We consider both simulated and real data for our ODT experiments.

\subsubsection{Simulated data} In our simulated setup, the test image (Figure \ref{fig:ODT_res_1}) that represents the RI contrast is a random sample from the dataset described in Section~\ref{sec:exp_aug_gans}: a disc with constant intensity $v$.

The measurements are simulated using the BPM of Section~\ref{sec:bpm} with an AWGN model of variance $\sigma_{\mathrm{n}}^{2} = 0.05$. We set the sampling steps to $\delta_{\mathrm{x}} = \delta_{\mathrm{y}} = \SI{0.1}{\micro\metre}$, the medium RI to $n_{\mathrm{b}}=1.52$, and the wavelength to $\lambda=\SI{0.406}{\micro\metre}$. We use $Q=20$ incident tilted plane waves with angles that are uniformly spaced in the range $[-\pi/12, \pi/12]$.

For this setting, we use the augmented WGAN prior of Section~\ref{sec:exp_aug_gans} in our reconstruction framework.

\subsubsection{Real data} In our experiment with real data, the sample is a 2D cross-section of two non-overlapping fibres immersed in oil ($n_{\mathrm{b}} = 1.525$) \cite{PhysRevApplied.9.034027}. The RI contrast of the sample is negative. A standard Mach-Zehnder interferometer relying on off-axis digital holography ($\lambda=\SI{0.450}{\micro\metre}$) is used to collect measurements from $Q=59$ views in the range $[-\pi/6, \pi/6]$.

We crop the acquired data such that the measurement vector for each view is of length $M' = 256$. We take the discretized region of interest to be of the size $(256 \times 256)$ and we set the sampling steps for BPM (used for reconstruction) to $\delta_{\mathrm{x}} = \delta_{\mathrm{y}} = \SI{0.1257}{\micro\metre}$. We assume an AWGN model of variance $\sigma_{\mathrm{n}}^{2} = 0.15$ for the measurements.

Here, the WGAN for our prior is trained on a synthetic dataset containing $100,\!000$ images of size $(256 \times 256)$, where each image consists of two non-overlapping discs with a constant intensity of one and a zero-valued background. The coordinates of the centers of the two discs are sampled from $U_{(20,235)}$ and their radii are sampled from $U_{[10,50]}$ subject to the constraint that they do not overlap. The distributions $p_{\mathrm{Z_1}}$ and $p_{\mathrm{Z_2}}$ are standard Gaussian distributions and the function $h$ is taken to be
\begin{equation}\label{eq:scaling_function_pr}
    h(x) = -\frac{0.1}{\sqrt{2 \pi}} \int_{-\infty}^{x} \mathrm{e}^{-\frac{t^2}{2}} \ud t. 
\end{equation}
The architectures for the generator and critic networks are detailed in Appendix \ref{app:architectures}. The WGAN is trained for $500$ epochs using RMSProp optimizers with a learning rate of $5 \times 10^{-5}$ and a batch size of $128$. The parameters $\lambda_{\text{gp}}$ and $n_{\text{critic}}$ are set as $10$ and $5$, respectively.

\subsubsection{Methods} For both settings, the estimate $\M s_{\text{init}}$ for MALA is obtained by the application of a filtered backpropagation algorithm that uses the Rytov approximation \cite{devaney1981inverse} to model the scattering. We collect $T$ samples from the posterior distribution using MALA with a step-size $\tau$ and burn-in period $T_{\mathrm{b}}$, and use them to compute the posterior mean and pixel-wise standard-deviation map. 

We compare our estimator with the TV-based method
\begin{align}\label{eq:odt_tv}
  \M s_{\text{TV}} = & \argmin_{\M s \in \R^K} \bigg( \sum_{q=1}^{Q} \|\M y_q - \M H_{\text{bpm}}(\M s; u^{\text{in}}_q)\|_{2}^{2} \nonumber \\ 
  & \mbox{ } + \ \tau \|\nabla \M s \|_{2,1} \ + \ \mathcal{I}(\M s) \bigg),
\end{align}
where $\mathcal{I}(\M s) = i_{+}(\M s)$ for the simulated data and $\mathcal{I}(\M s) = i_{-}(\M s)$ for the real data. This is a state-of-the-art method for ODT and is commonly used in practice \cite{kamilov2015learning, kamilov2016optical}. Moreover, it is well-suited for the constant-valued discs in our samples. The problem in \eqref{eq:odt_tv} is solved using FISTA initialized with $\M s_{\text{init}}$. The regularization parameter $\tau$ is tuned for optimal MSE performance in the simulated-data setting via a grid search, while it is tuned manually in the real-data setting.

\begin{figure*}[t]
  \centering
  \begin{tikzpicture}[xscale=5.5,yscale=2.2]
    \node (B) at (1,1) {\includegraphics[scale=0.45]{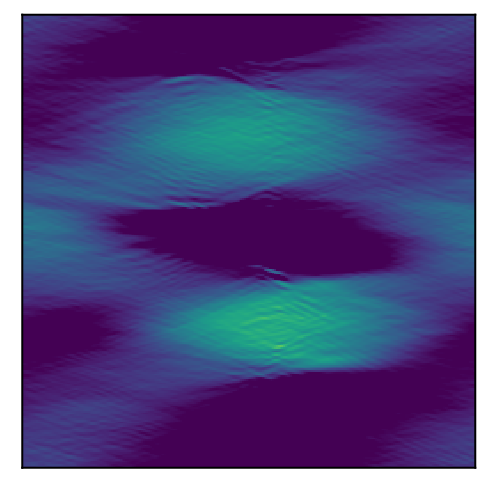}};
    \node at (B.north) {\small Initial reconstruction};
    \node (C) at (2,1) {\includegraphics[scale=0.45]{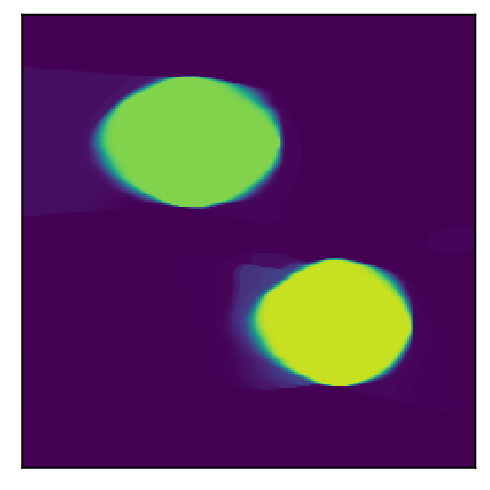}};
    \node at (C.north) {\small TV reconstruction};
    \node at (2.75,1) {\includegraphics[scale=0.45]{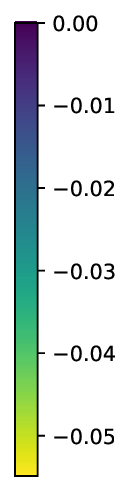}};
    \node (D) at (1,-1) {\includegraphics[scale=0.45]{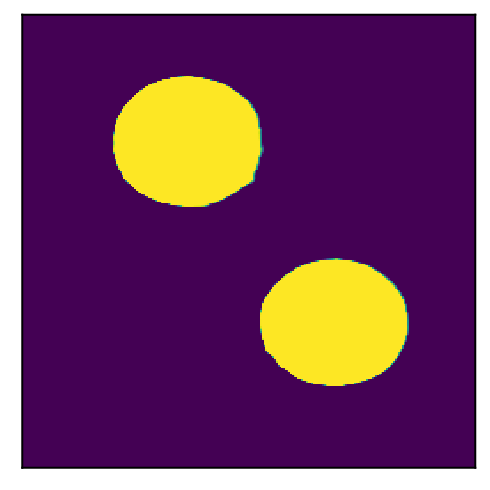}};
    \node at (D.north) {\small Posterior mean};
    \node (E) at (2,-1) {\includegraphics[scale=0.45]{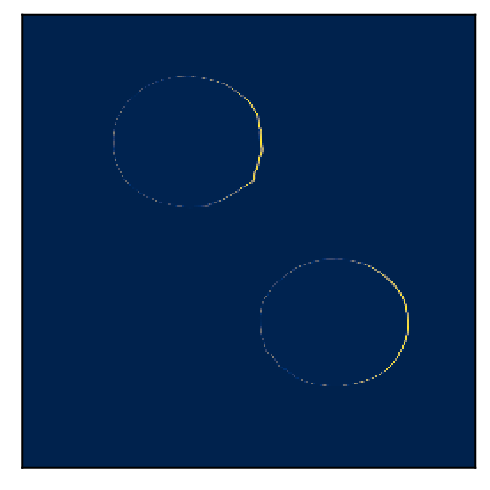}};
    \node at (E.north) {\small Posterior standard deviation};
    \node at (2.75,-1) {\includegraphics[scale=0.45]{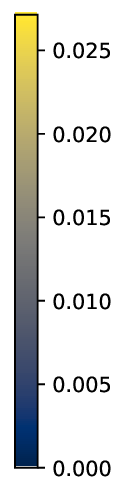}};
  \end{tikzpicture}
  \caption{Reconstructions for ODT (real data).}
  \label{fig:ODT_res_2}
\end{figure*}
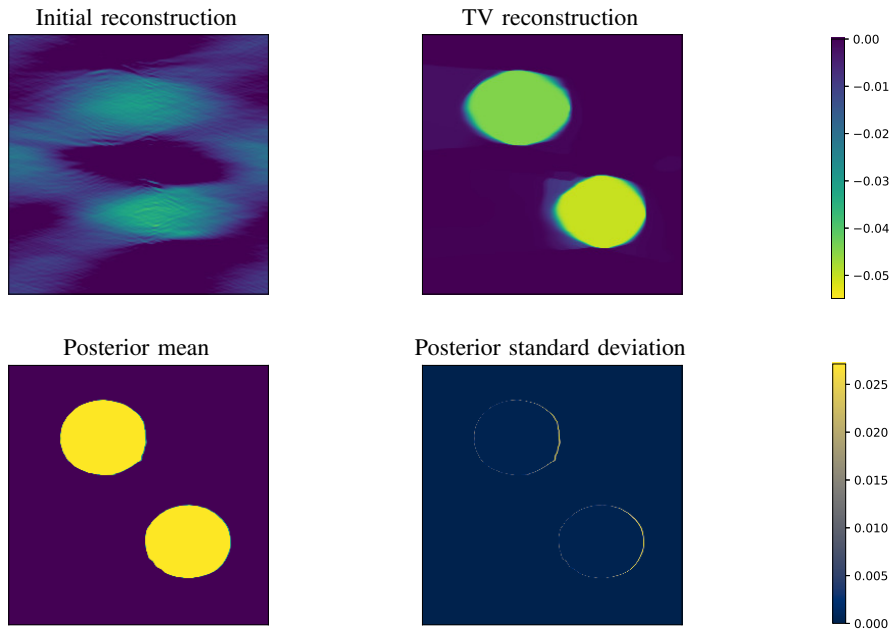

\subsubsection{Results} The settings that we consider for our ODT experiments are highly ill-posed as the incident waves only explore a limited range. As a result, the measurements lack information along the horizontal axis, which leads to the so-called missing-cone problem. For the first case (Figure \ref{fig:ODT_res_1}), we have that $v = 0.07, \eta = 2 \times 10^{-7}, T_{\mathrm{b}} = 2 \times 10^{4}$, and $T = 8 \times 10^{4}$. For the second case (Figure \ref{fig:ODT_res_2}), we have that $\eta = 5 \times 10^{-8}, T_{\mathrm{b}} = 15 \times 10^{4} $, and $T = 5 \times 10^{4}$.

In Figures \ref{fig:ODT_res_1} and \ref{fig:ODT_res_2}, we observe that the TV reconstructions (and the initial ones) are elongated in the horizontal direction due to the lack of information along this axis. However, the GAN-based estimator is able to overcome the missing-cone problem. It yields reconstructions whose quality is remarkable.

\subsection{Discussion}
With the help of the above-described experiments, we have demonstrated the potential of our deep-generative-prior-based Bayesian reconstruction framework for challenging nonlinear inverse problems. We now mention some directions for future work which can further improve this framework.

In the present form, our scheme lacks theoretical guarantees for MALA to be geometrically ergodic (convergence to the equilibrium distribution at a geometric rate). A topic of future work could be to investigate the imposition of appropriate constraints on the generative model such that the resulting posterior distribution satisfies certain smoothness and tail conditions \cite{durmus2022geometric} that ensure geometric ergodicity of MALA.

The performance of our scheme heavily relies on how well the prior models the object of interest. Thus, any progress on the side of designing and training high-quality large-scale deep generative models could be translated to our framework.

While the neural-network-like structure of our forward models make our approach tractable, like MCMC methods in general, it requires a lot of computation. It could be interesting to consider alternatives to MALA that might help in speeding up this approach.

\section{Conclusion}
We have presented a Bayesian reconstruction framework for nonlinear inverse problems where the prior information on the image of interest is encoded by a deep generative model. Specifically, we have designed a tractable posterior-sampling scheme based on the Metropolis-adjusted Langevin algorithm for the class of nonlinear inverse problems where the forward model has a neural-network-like computational structure. This class includes most practical imaging modalities. We have proposed the concept of augmented generative models. They allow us to tackle the problem of the quantitative recovery of images. Finally, we have illustrated the benefits of our framework by applying it to two nonlinear imaging modalities---phase retrieval and optical diffraction tomography.

\begin{appendices}

\section{Wasserstein Generative Adversarial Networks} \label{app:GANs}

Classical generative adversarial networks (GANs) \cite{goodfellow2014generative} are known to suffer from issues such as the instability of the training process \cite{salimans2016improved,arjovsky2017towards}, vanishing gradients, and mode collapse. The framework of Wasserstein GANs (WGANs) \cite{arjovsky2017wasserstein} is an alternative that alleviates these problems. 

Let $\mathcal{D}$ be a dataset consisting of samples drawn from a probability distribution $p_{\text{r}}$. The goal is to build a model using $\mathcal{D}$ that can generate samples that follow a distribution that closely approximates $p_{\text{r}}$. A WGAN consists of a generator network $\Op G_{\V \theta}: \R^d \rightarrow \R^K$ ($d \ll K$), where $\V \theta \in \R^{d_1}$ denotes its trainable parameters. It takes an input vector $\M z \in \R^d$, sampled from a fixed distribution $p_{\mathrm{Z}}$, and outputs $\Op G_{\V \theta}(\M z) \in \R^K$. The samples generated by this model follow some distribution $p_{\V \theta}$ that is characterized by $\Op G_{\V \theta}$ and $p_{\mathrm{Z}}$. Thus, the parameters $\V \theta$ need to be chosen such that $p_{\V \theta}$ approximates $p_{\text{r}}$ well. 

In the WGAN framework, the generator is trained to minimize the Wasserstein-1 (or Earth-Mover) distance between $p_{\text{r}}$ and $p_{\V \theta}$, which is given by
\begin{equation}\label{eq:W1_dist}
  W(p_{\text{r}}, p_{\V \theta}) = \inf_{\gamma \in \pi(p_{\text{r}}, p_{\V \theta})} \mathbb{E}_{(\M u, \M v) \sim \gamma}\big[\|\M u - \M v\|\big].
\end{equation} 
Here, $\pi(p_{\text{r}}, p_{\V \theta})$ is the collection of all joint distributions with marginals $p_{\text{r}}$ and $p_{\V \theta}$. The Kantorovich-Rubinstein duality theorem \cite{villani2009optimal} states that \eqref{eq:W1_dist} can be written as
\begin{equation}
  W(p_{\text{r}}, p_{\V \theta}) = \sup_{f \in \Spc X} \Big( \mathbb{E}_{\M u \sim p_{\text{r}}}[f(\M u)] - \mathbb{E}_{\M v \sim p_{\V \theta}}[f(\M v)] \Big),
\end{equation}
where $\Spc X = \{f:\R^K \rightarrow \R \ | \ f \text{ is 1-Lipschitz}\}$. The space $\Spc X$ is then replaced by a family of $1$-Lipschitz functions represented by a critic neural network $\Op D_{\V \phi}: \R^K \rightarrow \R$ with appropriately constrained parameters $\V \phi \in \R^{d_2}$. This leads to the minimax problem
\begin{equation}\label{eq:minimax}
  \min_{\V \theta \in \R^{d_1}} \max_{\V \phi \in \Spc Y} \Big( \mathbb{E}_{\M u \sim p_{\text{r}}}[\Op D_{\V \phi}(\M u)] - \mathbb{E}_{\M v \sim p_{\V \theta}}[\Op D_{\V \phi}(\M v)] \Big),
\end{equation}
where $\Spc Y = \{\V \phi \in \R^{d_2} \ | \ \Op D_{\V \phi} \text{ is 1-Lipschitz}\}$. In \cite{arjovsky2017wasserstein}, the authors enforce the $1$-Lipschitz condition on $\Op D_{\V \phi}$ by clipping its weights during training.
Instead, the $1$-Lipschitz constraint can also be enforced by adding a gradient penalty to the cost function in \eqref{eq:minimax} \cite{gulrajani2017improved}. The regularized minimax problem becomes
\begin{align}\label{eq:minimax_gp}
  &\min_{\V \theta \in \R^{d_1}} \max_{\V \phi \in \R^{d_2}} \ \Big(\mathbb{E}_{\M u \sim p_{\text{r}}}[\Op D_{\V \phi}(\M u)] - \mathbb{E}_{\M v \sim p_{\V \theta}}[\Op D_{\V \phi}(\M v)\big] \nonumber \\
  & \mbox{ } +\lambda_{\text{gp}} \mathbb{E}_{\M w \sim p_{\text{int}}}\big[ (\|\nabla_{\M w} \Op D_{\M \phi}(\M w) \| - 1)^2 \big] \Big),
\end{align}
where a point $\M w \sim p_{\text{int}}$ is obtained by sampling uniformly along straight lines between points drawn from $p_{\text{r}}$ and $p_{\V \theta}$, and $\lambda_{\text{gp}} > 0$ is a hyperparameter. In practice, Problem \eqref{eq:minimax_gp} is solved using mini-batch stochastic-gradient algorithms in an alternating manner. During each iteration for the critic, we collect a batch of samples $\{\M x_n\}_{n=1}^{N_c}$ from the dataset $\Spc D$. We sample vectors $\{\M z_n\}_{n=1}^{N_c}$ from $p_{\mathrm{Z}}$ and a sequence of numbers $\{\alpha_n\}_{n=1}^{N_c}$ from the uniform distribution $U_{[0,1]}$, and we construct $\M w_n = \alpha_n \M x_n + (1 - \alpha_n) \Op G_{\V \theta}(\M z_n)$. The critic parameters are then updated by ascending along the gradient given by
\begin{align}
&\frac{1}{N_c} \nabla_{\V \phi} \Bigg(\sum_{n=1}^{N_c} \Op D_{\V \phi}(\M x_n) - \Op D_{\V \phi}(\Op G_{\V \theta}(\M z_n)) \nonumber \\ 
& \mbox{ } + \lambda_{\text{gp}} (\|\nabla_{\M w_n} \Op D_{\V \phi}(\M w_n) \| - 1)^2 \Bigg).  
\end{align}   
During each iteration for the generator, we sample latent vectors $\{\M z_n\}_{n=1}^{N_g}$ from $p_{\mathrm{Z}}$. The generator parameters are then updated by descending along the gradient given by  
\begin{equation}
  \frac{1}{N_g} \nabla_{\V \theta} \Bigg(\sum_{n=1}^{N_g} -\Op D_{\V \phi}(\Op G_{\V \theta}(\M z_n)) \Bigg).
\end{equation}
Typically, for every generator iteration, the critic is trained for $n_{\text{critic}}$ iterations.  

\section{Properties of the Posterior Distribution} \label{app:prop_post}

\subsection{Well-posedness}

A Bayesian inverse problem is said to be well-posed in some metric on the space of probability measures if the posterior distribution exists, is unique, and is continuous with respect to the measurements for the chosen metric \cite{latz2020well}. Here, we present sufficient conditions from \cite[Assumptions 3.5, 3.10 and Theorems 3.6, 3.12]{latz2020well} that guarantee the well-posedness of our problem in the latent space, that is, with respect to $p_{\mathrm{Z}|\mathrm{Y}}$ as described in Section \ref{sec:post}.

The following conditions are stated for $p_{\mathrm{Z}}$-almost every (a.e.) $\M z \in \R^{d+1}$ and every $\M y \in \R^M$.

\noindent \textbf{Conditions.}
\begin{enumerate}
    \item $p_{\mathrm{Y}|\mathrm{Z}}(\cdot | \M z)$ is a strictly positive pdf.
    \item $\int_{\R^{d+1}} |p_{\mathrm{Y}|\mathrm{Z}}(\M y | \M z')| p_{\mathrm{Z}}(\M z') \ud \M z' < \infty $
    \item There exists $g$ with $\int_{\R^{d+1}} |g(\M z')| p_{\mathrm{Z}}(\M z') \ud \M z' < \infty $ such that $p_{\mathrm{Y}|\mathrm{Z}}(\M y' | \cdot) \leq g$ for all $\M y' \in \R^{M}$.
    \item $p_{\mathrm{Y}|\mathrm{Z}}(\cdot | \M z)$ is continuous.
    \item There exists $g'$ with $\int_{\R^{d+1}} |g'(\M z')| p_{\mathrm{Z}}(\M z') \ud \M z' < \infty $ such that $\|\M z''\|_{2}^{p} \ p_{\mathrm{Y}|\mathrm{Z}}(\M y' | \M z'') \leq g'(\M z'')$, where $p \in [1, \infty)$, for $p_{\mathrm{Z}}$-a.e. $\M z'' \in \R^{d+1}$ and all $\M y' \in \R^{M}$. 
\end{enumerate}
If the conditions $(1)-(4)$ hold, our Bayesian inverse problem in the latent space is well-posed in the Prokhorov, Hellinger and total-variation distances. In addition, if condition $(5)$ holds, then the problem is also well-posed in the Wasserstein $p$-distance.

For additive white-Gaussian-noise (AWGN) models, the conditions $(1)-(4)$ are satisfied for any physical forward model $\M H$ and prior distribution $p_{\mathrm{Z}}$. Further, if $p_{\mathrm{Z}}$ is such that $\int \|\M z'\|_{2}^{p} \ p_{\mathrm{Z}}(\M z') \ud \M z' < \infty $ (\textit{e.g.,} Gaussian distribution), condition $(5)$ is also satisfied \cite[Corollary 5.1]{latz2020well}. As for the Poisson-noise models used in some of our experiments, they do not fall within this framework of well-posedness developed in \cite{latz2020well}.

\subsection{Existence of Moments}
Based on Proposition $3.6$ in \cite{holden2022bayesian}, we also present some conditions under which the moments of our posterior distribution $p_{\mathrm{S}|\mathrm{Y}}$ exist. If the augmented deep generative prior $\Op G_{h}$ is Lipschitz-continuous and the prior distribution $p_{\mathrm{Z}}$ has finite moments $\mathbb{E}_{p_{\mathrm{Z}}}[|\M z|^{k}]$ for $k=1, 2, \ldots, K$, then the $K$th posterior moment $\mathbb{E}_{p_{\mathrm{S}|\mathrm{Y}}}[|\M s|^{K}]$ exists for almost all measurements $\M y$.

The typical choice for $p_{\mathrm{Z}}$ is the standard Gaussian distribution, which has finite moments. The Lipschitz-continuity of $\Op G_h$ is guaranteed if the generative network $\Op G$ and the function $h$ are both Lipschitz-continuous and bounded. The Lipschitz condition on the network $\Op G$ holds when its weights and biases are finite-valued and it consists of Lipschitz-continuous activation functions (\textit{e.g.,} ReLU, sigmoid). The boundedness of $\Op G$ is ensured when the activation function in the output layer is bounded (such as the sigmoid function). These are conditions that are satisfied by the networks used in Section \ref{sec:experiments}. Further, in our experiments, we choose the function $h$ to be a scaled version of the cumulative density function of the standard normal distribution, which is Lipschitz-continuous and bounded.

\begin{table}[t]
  \begin{subtable}[t]{\linewidth}
      \centering
      \begin{tabular}{ c | c }
      \hline\hline
      Layers & Output shape \\
      \hline
      Conv $4 \times 4$ + LReLU & $512 \times 4 \times 4$\\
      Conv $3 \times 3$ + LReLU & $512 \times 4 \times 4$\\
      \hline
      Upsample & $512 \times 8 \times 8$ \\
      Conv $3 \times 3$ + LReLU & $256 \times 8 \times 8$\\
      \hline
      Upsample & $256 \times 16 \times 16$ \\
      Conv $3 \times 3$ + LReLU & $128 \times 16 \times 16$\\
      \hline
      Upsample & $128 \times 32 \times 32$ \\
      Conv $3 \times 3$ + LReLU & $64 \times 32 \times 32$\\
      \hline
      Upsample & $64 \times 64 \times 64$ \\
      Conv $3 \times 3$ + LReLU & $32 \times 64 \times 64$\\
      \hline
      Upsample & $32 \times 128 \times 128$ \\
      Conv $3 \times 3$ + LReLU & $16 \times 128 \times 128$\\
      \hline
      Conv $1 \times 1$ + Sigmoid & $1 \times 128 \times 128$\\
      \hline\hline
     \end{tabular}
     \caption{Generator network with $(128 \times 1 \times 1)$ input shape.}
     \label{tab:generator_network_beads}
     \vspace{0.5cm}
     \begin{tabular}{ c | c }
      \hline\hline
      Layers & Output shape \\
      \hline
      Conv $1 \times 1$ + LReLU & $16 \times 128 \times 128$\\
      Conv $3 \times 3$ + LReLU & $16 \times 128 \times 128$\\
      Conv $3 \times 3$ + LReLU & $32 \times 128 \times 128$\\
      Downsample & $32 \times 64 \times 64$\\
      \hline
      Conv $3 \times 3$ + LReLU & $64 \times 64 \times 64$\\
      Downsample & $64 \times 32 \times 32$\\
      \hline
      Conv $3 \times 3$ + LReLU & $128 \times 32 \times 32$\\
      Downsample & $128 \times 16 \times 16$\\
      \hline
      Conv $3 \times 3$ + LReLU & $256 \times 16 \times 16$\\
      Downsample & $256 \times 8 \times 8$\\
      \hline
      Conv $3 \times 3$ + LReLU & $512 \times 8 \times 8$\\
      Downsample & $512 \times 4 \times 4$\\
      \hline
      Conv $3 \times 3$ + LReLU & $512 \times 4 \times 4$\\
      Conv $4 \times 4$ + LReLU & $512 \times 1 \times 1$\\
      \hline
      Reshape & $1 \times 512$ \\
      Fully-connected & $1 \times 1$\\
      \hline\hline
     \end{tabular}
     \caption{Critic network with $(1 \times 128 \times 128)$ input shape.}
     \label{tab:critic_network_beads}
   \end{subtable}
   \caption{Generator and critic architectures (single disc). The negative slope for LReLU is set as $0.2$. The upsampling layer uses nearest-neighbor interpolation while the downsampling layer involves max pooling.}
   \label{tab:GAN_architecture_beads}
\end{table}

\section{WGAN Architectures} \label{app:architectures}
The generator and critic architectures used for datasets consisting of constant-valued discs are shown in Table \ref{tab:GAN_architecture_beads} and \ref{tab:GAN_architecture_two_beads}. The architectures used for the MNIST and Fashion MNIST datasets are shown in Table \ref{tab:GAN_architecture_mnist} and \ref{tab:GAN_architecture_fmnist}, respectively.

\vspace{1cm}

\begin{table}[!h]
  \begin{subtable}[t]{\linewidth}
      \centering
      \begin{tabular}{c | c }
      \hline\hline
      Layers & Output shape \\
      \hline
      Fully-connected + LReLU & $1 \times 128$\\
      Fully-connected + Batch-norm + LReLU & $1 \times 256$\\
      Fully-connected + Batch-norm + LReLU & $1 \times 512$\\
      Fully-connected + Batch-norm + LReLU & $1 \times 1024$\\
      Fully-connected + Sigmoid & $1 \times 784$\\
      \hline\hline
     \end{tabular}
     \caption{Generator network with $(1 \times 100)$ input shape.}
     \label{tab:generator_network_mnist}
     \vspace{0.5cm}
     \begin{tabular}{c | c }
      \hline\hline
      Layers & Output shape \\ 
      \hline
      Fully-connected + LReLU & $1 \times 512$\\
      Fully-connected + LReLU & $1 \times 256$\\
      Fully-connected  & $1 \times 1$\\
      \hline\hline
     \end{tabular}
     \caption{Critic network with $(1 \times 784)$ input shape.}
     \label{tab:critic_network_mnist}
   \end{subtable}
   \caption{Generator and critic architectures (MNIST). The negative slope for LReLU is set as $0.2$.}
   \label{tab:GAN_architecture_mnist}
\end{table}

\vspace{0.5cm}

\begin{table}[!h]
  \begin{subtable}[!h]{\linewidth}
      \centering
      \begin{tabular}{c | c }
      \hline\hline
      Layers & Output shape \\
      \hline
      Fully-connected + Batch-norm + ReLU & $1 \times 1024$\\
      Fully-connected + Batch-norm + ReLU & $1 \times 6272$\\
      Reshape & $128 \times 7 \times 7$\\
      ConvTranspose $4 \times 4$ + Batch-norm + ReLU & $64 \times 14 \times 14$ \\
      ConvTranspose $4 \times 4$ + Sigmoid & $1 \times 28 \times 28$ \\
      \hline\hline
     \end{tabular}
     \caption{Generator network with $(1 \times 100)$ input shape.}
     \label{tab:generator_network_fmnist}
     \vspace{0.5cm}
     \begin{tabular}{c | c }
      \hline\hline
      Layers & Output shape \\ 
      \hline
      Conv $4 \times 4$ + LReLU & $64 \times 14 \times 14$\\
      Conv $4 \times 4$ + Batch-norm + LReLU & $128 \times 7 \times 7$\\
      Reshape & $1 \times 6272$ \\
      Fully-connected + Batch-norm + LReLU & $1 \times 1024$\\
      Fully-connected  & $1 \times 1$\\
      \hline\hline
     \end{tabular}
     \caption{Critic network with $(1 \times 28 \times 28)$ input shape.}
     \label{tab:critic_network_fmnist}
   \end{subtable}
   \caption{Generator and critic architectures (Fashion-MNIST). The negative slope for LReLU is set as $0.2$.}
   \label{tab:GAN_architecture_fmnist}
\end{table}

\begin{table}[t]
  \begin{subtable}[t]{\linewidth}
      \centering
      \begin{tabular}{ c | c }
      \hline\hline
      Layers & Output shape \\
      \hline
      Conv $4 \times 4$ + LReLU & $256 \times 4 \times 4$\\
      Conv $3 \times 3$ + LReLU & $256 \times 4 \times 4$\\
      \hline
      Upsample & $256 \times 8 \times 8$ \\
      Conv $3 \times 3$ + LReLU & $128 \times 8 \times 8$\\
      \hline
      Upsample & $128 \times 16 \times 16$ \\
      Conv $3 \times 3$ + LReLU & $64 \times 16 \times 16$\\
      \hline
      Upsample & $64 \times 32 \times 32$ \\
      Conv $3 \times 3$ + LReLU & $32 \times 32 \times 32$\\
      \hline
      Upsample & $32 \times 64 \times 64$ \\
      Conv $3 \times 3$ + LReLU & $16 \times 64 \times 64$\\
      \hline
      Upsample & $16 \times 128 \times 128$ \\
      Conv $3 \times 3$ + LReLU & $8 \times 128 \times 128$\\
      \hline
      Upsample & $8 \times 256 \times 256$ \\
      Conv $3 \times 3$ + LReLU & $4 \times 256 \times 256$\\
      \hline
      Conv $1 \times 1$ + Sigmoid & $1 \times 256 \times 256$\\
      \hline\hline
     \end{tabular}
     \caption{Generator network with $(128 \times 1 \times 1)$ input shape.}
     \label{tab:generator_network_two_beads}
     \vspace{0.5cm}
     \begin{tabular}{ c | c }
      \hline\hline
      Layers & Output shape \\
      \hline
      Conv $1 \times 1$ + LReLU & $4 \times 256 \times 256$\\
      Conv $3 \times 3$ + LReLU & $4 \times 256 \times 256$\\
      Conv $3 \times 3$ + LReLU & $8 \times 256 \times 256$\\
      Downsample & $8 \times 128 \times 128$\\
      \hline
      Conv $3 \times 3$ + LReLU & $16 \times 128 \times 128$\\
      Downsample & $16 \times 64 \times 64$\\
      \hline
      Conv $3 \times 3$ + LReLU & $32 \times 64 \times 64$\\
      Downsample & $32 \times 32 \times 32$\\
      \hline
      Conv $3 \times 3$ + LReLU & $64 \times 32 \times 32$\\
      Downsample & $64 \times 16 \times 16$\\
      \hline
      Conv $3 \times 3$ + LReLU & $128 \times 16 \times 16$\\
      Downsample & $128 \times 8 \times 8$\\
      \hline
      Conv $3 \times 3$ + LReLU & $256 \times 8 \times 8$\\
      Downsample & $256 \times 4 \times 4$\\
      \hline
      Conv $3 \times 3$ + LReLU & $256 \times 4 \times 4$\\
      Conv $4 \times 4$ + LReLU & $256 \times 1 \times 1$\\
      \hline
      Reshape & $1 \times 256$\\
      Fully-connected & $1 \times 1$\\
      \hline\hline
     \end{tabular}
     \caption{Critic network with $(1 \times 256 \times 256)$ input shape.}
     \label{tab:critic_network_two_beads}
   \end{subtable}
   \caption{Generator and critic architectures (two non-overlapping discs). The negative slope for LReLU is set as $0.2$. The upsampling layer uses nearest-neighbor interpolation while the downsampling layer involves max pooling.}
   \label{tab:GAN_architecture_two_beads}
\end{table}

\end{appendices}

\section*{Acknowledgments}
We would like to thank Dr. Joowon Lim and Prof. Demetri Psaltis for providing us with real data for optical diffraction tomography, and Dr. Aleix Boquet-Pujadas for helpful discussions.

\bibliographystyle{IEEEtran.bst} 
\bibliography{refs}
\end{document}